\documentclass[10pt,twocolumn,letterpaper]{article}

\usepackage[pagenumbers]{cvpr} 

\usepackage{graphicx}
\usepackage{amsmath}
\usepackage{amssymb}
\usepackage{booktabs}
\usepackage{multirow}
\usepackage{color, colortbl}
\usepackage{enumitem}
\usepackage{wrapfig}
\usepackage[dvipsnames]{xcolor}
\usepackage[symbol]{footmisc}

\makeatletter
\@namedef{ver@everyshi.sty}{}
\makeatother
\usepackage{tikz}
\usepackage[pagebackref,breaklinks,colorlinks]{hyperref}

\usepackage[capitalize]{cleveref}
\crefname{section}{Sec.}{Secs.}
\Crefname{section}{Section}{Sections}
\Crefname{table}{Table}{Tables}
\crefname{table}{Tab.}{Tabs.}

\def\cvprPaperID{921} 
\def\confName{CVPR}
\def\confYear{2022}

\newcommand{\StatesSet}{\mathcal{S}}
\newcommand{\ActionsSet}{\mathcal{A}}
\newcommand{\TransitionSet}{\mathcal{T}}

\newcommand{\discountRate}{\gamma}

\newcommand{\mdp}{\mathcal{M}}
\newcommand{\taskname}{dynamic grasp synthesis}
\newcommand{\methodname}{\textit{D-Grasp}\xspace}
\newcommand{\supmat}{supp.~material\xspace}
\newcommand{\raisim}{RaiSim\xspace}

\newcommand{\greycellcol}{\cellcolor[HTML]{E4E4E4}}

\definecolor{blueish}{rgb}{0.0, 0.3, .6}

\newcommand{\figref}[1]{Fig.~\ref{#1}}
\newcommand{\secref}[1]{Section~\ref{#1}}

\newcommand{\tabref}[1]{Tab.~\ref{#1}}
\newcommand{\update}[1]{{\color{black}{#1}}}
\newcommand{\camera}[1]{{\color{black}{#1}}}
\newcommand{\Fig}[1]{Fig.~\ref{fig:#1}}

\newcommand{\Tab}[1]{Tab.~\ref{tab:#1}}

\definecolor{Gray}{gray}{0.9}

\definecolor{hexorange}{HTML}{FF9900}
\definecolor{hexblue}{HTML}{3C78D8}
\DeclareRobustCommand*\circledorange[1]{\tikz[baseline=(char.base)]{
            \node[hexorange,shape=circle,draw,inner sep=1pt, thick=8pt, scale=0.8] (char) {#1};}}
       
\DeclareRobustCommand*\circledblue[1]{\tikz[baseline=(char.base)]{
            \node[hexblue,shape=circle,draw,inner sep=1pt, thick=8pt, scale=0.8] (char) {#1};}}

\newif\ifshowcomments

\ifshowcomments
        \newcommand{\note}[3]{{\textcolor{#2}{[#1: #3]}}}
        \newcommand{\OH}[1]{\note{OH}{red}{#1}}
		\newcommand{\EA}[1]{\note{EA}{green}{#1}}
		\newcommand{\mkocabas}[1]{\note{MK}{cyan}{#1}}
		\newcommand{\JS}[1]{\note{JS}{brown}{#1}}
		\newcommand{\JH}[1]{\note{JH}{magenta}{#1}}
		\newcommand{\SC}[1]{\note{SC}{teal}{#1}}
		\newcommand{\AS}[1]{\note{AS}{olive}{#1}}
		\newcommand{\MK}[1]{\note{MKA}{blue}{#1}}
		
\else
        \newcommand{\note}[3]{\unskip}
        \newcommand{\OH}[1]{\unskip}
		\newcommand{\EA}[1]{\unskip}
		\newcommand{\mkocabas}[1]{\unskip}
		\newcommand{\JS}[1]{\unskip}
		\newcommand{\JH}[1]{\unskip}
		\newcommand{\SC}[1]{\unskip}
		\newcommand{\AS}[1]{\unskip}
		\newcommand{\MK}[1]{\unskip}
\fi

\newcommand{\oh}[1]{\OH{#1}}
\newcommand{\jh}[1]{\JH{#1}}

\newcommand{\posevec}{\mathbf{q}_h}
\newcommand{\posevecgen}{\mathbf{q}}

\newcommand{\posvec}{\mathbf{x}}
\newcommand{\goalvec}{\mathbf{g}}
\newcommand{\vertexvec}{\mathbf{v}}
\newcommand{\posesix}{\mathbf{T}}
\newcommand{\grasplabel}{\mathbf{D}}
\newcommand{\goals}{\mathbf{G}}
\newcommand{\forcevec}{\mathbf{f}}

\newcommand{\actions}{\mathbf{a}}
\newcommand{\statevec}{\mathbf{s}}
\newcommand{\torques}{\boldsymbol{\tau}}
\newcommand{\policyvec}{\boldsymbol{\pi}}
\newcommand{\gpolicyvec}{\boldsymbol{\pi}_{g}}
\newcommand{\mpolicyvec}{\boldsymbol{\pi}_{m}}

\usepackage{etoolbox}
\patchcmd{\thebibliography}{\section*{\refname}}{}{}{}

\renewcommand{\thefootnote}{\fnsymbol{footnote}}

\begin{document}

\title{DynGrasp: Physically Plausible Hand-Object Interaction Synthesis}

\title{D-Grasp: Physically Plausible Dynamic Grasp Synthesis\\ for Hand-Object Interactions}

\author{%
  Sammy Christen$^{1}$\quad \;Muhammed Kocabas$^{1,2}$\quad \; Emre Aksan$^1$\quad \ \\ Jemin Hwangbo$^3$\quad \; Jie Song$^{1\dag}$ \quad \; Otmar Hilliges$^1$\quad \\
  \normalsize $^1$Department of Computer Science, ETH Zurich \quad
  \normalsize $^2$Max Planck Institute for Intelligent Systems, T\"{u}bingen \quad \\
  \normalsize $^3$ Department of Mechanical Engineering, KAIST\quad \\
}

\twocolumn[{%
\renewcommand\twocolumn[1][]{#1}%
\maketitle

\vspace{-2.5em}
\begin{center}
    \captionsetup{type=figure}
   \includegraphics[width=1.0\linewidth]{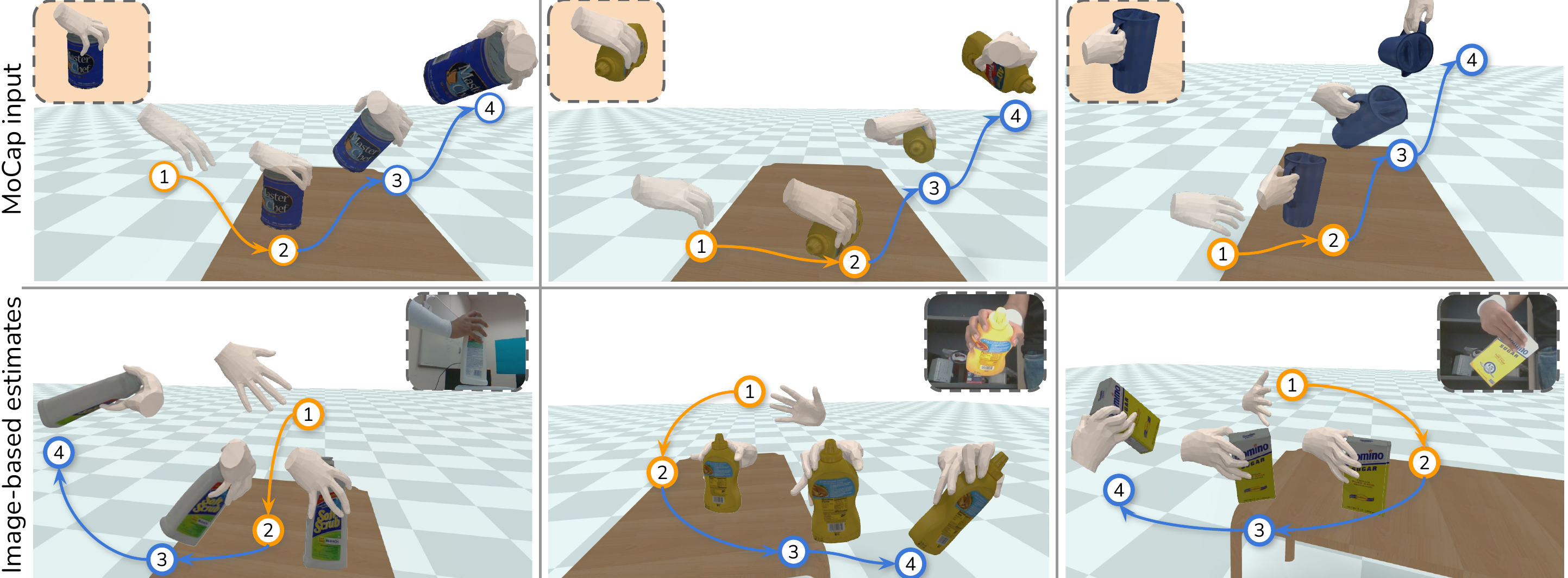}
   \vspace{-0.6cm}
    \captionof{figure}{\textbf{Dynamic Grasp Synthesis}: 
    Our method learns diverse grasps from static grasp labels (shown in insets), originating from existing datasets, grasp synthesis or image-based estimates. Our approach can then synthesize diverse dynamic sequences with the objects in-hand. We decompose the task into: stable grasping \circledorange{\textcolor{black}{1}}-\circledorange{\textcolor{black}{2}}, followed by the synthesis of a 3D global motion to move the object into a 6D target pose \circledblue{\textcolor{black}{3}}-\circledblue{\textcolor{black}{4}}. The hand-pose is continuously adjusted to ensure a stable grasp, leading to physically plausible and human-like sequences.}
\label{fig:teaser}
\end{center}%
}]

\begin{abstract}
\vspace{-1em}
We introduce the \taskname~task: given an object with a known 6D pose and a grasp reference, our goal is to generate motions that move the object to a target 6D pose. 
%
This is challenging, because it requires reasoning about the complex articulation of the human hand and the intricate physical interaction with the object. 
We propose a novel method that frames this problem in the reinforcement learning framework and leverages a physics simulation, both to learn and to evaluate such dynamic interactions. 
A hierarchical approach decomposes the task into low-level grasping and high-level motion synthesis. 
It can be used to generate novel hand sequences that approach, grasp, and move an object to a desired location, while retaining human-likeness. 
We show that our approach leads to stable grasps and generates a wide range of motions. 
Furthermore, even imperfect labels can be corrected by our method to generate dynamic interaction sequences. Video and code are available at: \footnotesize{\url{https://eth-ait.github.io/d-grasp/}}.
\vspace{-0.8cm}
\end{abstract}

\def\thefootnote{\dag}\footnotetext{Corresponding author}

\section{Introduction}

A key problem in computer vision is to understand how humans interact with their surroundings. Because hands are our primary means of manipulation with the physical world, there has been an intense interest in hand-object pose estimation \cite{tekin2019h+, hasson2019learning, hasson2020leveraging, chao2021dexycb, hampali2020ho3d, taheri2020grabnet, jiang2021graspTTA} and the \emph{synthesis} of \emph{static} grasps for a given object \cite{jiang2021graspTTA, karunratanakul2021skeletondriven, li2021artiboost, taheri2020grabnet}. 
However, human grasping is not limited to a single time instance, but involves a continuous interaction with objects in order to \emph{move} them. It requires maintaining a stable grasp throughout the interaction, introducing intricate dynamics to the task.
This involves reasoning about the complex physical interactions between the dexterous hand and the manipulated object, including collisions, friction, and dynamics.
A generative model that can synthesize realistic and physically plausible object manipulation sequences would have many downstream applications in AR/VR, robotics and HCI. 

We propose the new task of \textit{\taskname}. Given an object with a known 6D pose and a static grasp reference, our goal is to generate a grasping motion and to move the object to a target 6D pose in a natural and physically-plausible way. 
This new setting adds several challenges. 
First, the object geometry and the spatial configuration of the object and the hand need to be considered in continuous interaction.
%
Second, contacts between the hand and object are crucial in maintaining stability of the grasps, where even a small error in hand pose may lead to an object slipping. Moreover, contact is typically unobservable in images \cite{ehsani2020force} and measuring the stability of a grasp is very challenging in a static setting. 
Finally, synthesizing sequences of hand motion requires the generation of smooth and plausible trajectories. While prior work investigates the control of dexterous hands by learning from full demonstration trajectories \cite{garciahernando2020iros,rajeswaran2017rss}, we address the generation of hand motion from only a single-frame grasp reference. This is a more challenging setting, because the generation of human-like hand-object interaction trajectories without dense supervision is not straightforward.

\update{Taking a step towards this goal, we propose \methodname, which generates physically plausible grasping motions with only a single grasp reference as input (\Fig{teaser}).}
Concretely, we formulate the \textit{\taskname} task as a reinforcement learning (RL) problem and propose a policy learning approach that leverages a physics simulation.
Our RL-based approach considers the underlying physical phenomena and compensates data scarcity via exploration in the physics simulation. 
This ensures physical plausibility, e.g., there is no hand-object interpenetration and the fingers exert enough force on the object to hold it without slipping. 






Specifically, we introduce a hierarchical framework that consists of a low-level grasping policy and a high-level motion synthesis module. The grasping policy's purpose is to establish and maintain a stable grasp, whereas the motion synthesis module generates a motion to move the object to a user-specified target position.
To guide the low-level grasping policy, we require a single grasp label corresponding to a static hand pose, which can be obtained either from a hand-grasping dataset~\cite{chao2021dexycb,hampali2020ho3d}, a state-of-the-art grasp synthesis method~\cite{jiang2021graspTTA} \camera{or via an image-based pose estimator \cite{grady2021contactopt}}. 
Crucially, we propose a reward function that is parameterized by the grasp label to incentivize the fingers to reach contact points on the object, leading to human-like grasps. 
Our high-level motion synthesis module generates motions that move the hand and object to the final target pose. Importantly, the low-level policy continually controls the grasp to not drop the object. 

In our experiments, we first demonstrate that samples from motion capture, static grasp synthesis or image-based pose estimates often do not lead to stable grasps when evaluated in a physics  simulation (\figref{fig:grasp_qual}). 
\oh{Somewhere (not here, maybe in\figref{fig:grasp_qual} we should explain that this is expected: observing contact is hard if not impossible and therefore GT annotations are never really GT.}
We then present how our method can learn to produce physically plausible and stable grasps when guided by such labels. Next, we set out to generate motions with the object in-hand to reach a wide range of target poses. We provide an extensive ablation,  revealing the importance of the hierarchical approach and the reward formulation for \taskname.


    

Our contributions can be summarized as follows: i) We introduce the new task of \emph{\taskname}. ii) We propose \methodname, an RL-based method to synthesize physically-plausible and natural hand-object interactions. iii) We show that our method can generate grasp motions with static grasp references, \camera{which can originate from motion capture, static grasp synthesis or image-based pose estimation} 
\noindent. We will release our code for research purposes.


\section{Related Work}
\begin{figure*}[t]
\begin{center}
   \includegraphics[width=0.98\textwidth]{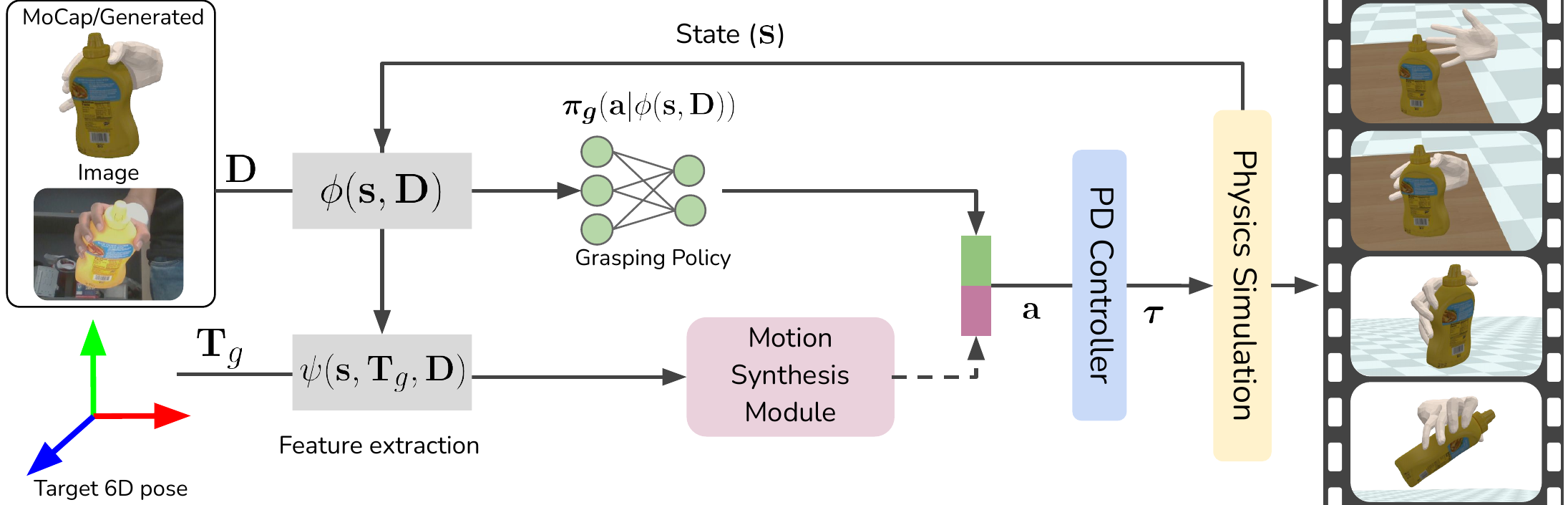}
\end{center}
\vspace{-0.4cm}
   \caption{\textbf{Method Overview:} Taking a single, static grasp label  $\grasplabel$ and a target object 6D pose $\mathbf{T}_g$ as input (leftmost), \methodname produces sequences of dynamic hand-object interactions (rightmost). To do so, we propose a hierarchical framework that consists of a low-level grasping policy $\boldsymbol{\pi}_g(\cdot)$ and a high-level motion synthesis module. In the \emph{grasping} phase, only the grasping policy is active and finds a stable grasp on the object. In the subsequent \emph{motion synthesis} phase, both the grasping policy and the motion synthesis module act concurrently. The actions $\mathbf{a}$ consist of joint targets. These are combined and passed to a PD-controller that computes the required torques $\boldsymbol{\tau}$ to control a MANO-based hand model in a physics simulation. The physics simulation updates the state $\mathbf{s}$ which serves as input to a reward formulation (\secref{method_reward}) that forms our supervision signal and incentivizes the hand to approach and grasp the object and to move it to the target 6D pose. We introduce two feature extraction layers ($\phi(\cdot)$ an $\psi(\cdot)$) that utilize the environment state $\mathbf{s}$ and grasp label $\grasplabel$ to find a suitable representation for the grasping policy and the motion synthesis module.}
\vspace{-0.2cm}
\label{fig:method_overview}
\end{figure*}

\textbf{Human Grasp Prediction}
Recently, hand-object interaction has received much research attention. This growth is accelerated by the introduction of datasets that contain both hand and object annotations \cite{brahmbhatt2019contactdb, brahmbhatt2020contactpose, corona2020ganhand, taheri2020grabnet, hampali2020ho3d, chao2021dexycb, kwon2021h2o}. 
Leveraging this data, a large number of methods attempt to estimate grasp parameters, such as the hand and object pose, directly from RGB images \cite{karunratanakul2020grasping, yang2021cpf, doosti2020hope, tekin2019h+, hasson2019learning, hasson2020leveraging, liu2021semi, cao2021reconstructing}. Some predict the mesh of the hand and the object directly \cite{hasson2019learning}, or assume a known object and predict its 6DoF in addition to the hand \cite{hasson2020leveraging, cao2021reconstructing, yang2021cpf, liu2021semi}. Others predict 3D keypoints and 6 DoF pose of the object \cite{doosti2020hope, tekin2019h+} or produce an implicit surface representation of the grasping hands \cite{karunratanakul2020grasping}. To improve the prediction accuracy of the grasp, many of these works incorporate additional contact losses \cite{karunratanakul2020grasping, hasson2019learning} or propose a contact-aware refinement step \cite{yang2021cpf, cao2021reconstructing}. 
More directly related are methods that attempt to generate static grasps given an object and sometimes also  information about the hand \cite{brahmbhatt2019contactdb, brahmbhatt2020contactpose, karunratanakul2021skeletondriven, zhu2021toward, jiang2021graspTTA, karunratanakul2020grasping, taheri2020grabnet}. Generally, these approaches either predict a contact map on the object \cite{brahmbhatt2019contactdb, brahmbhatt2020contactpose, jiang2021graspTTA} or synthesize the joint-angle configuration of the grasping hand \cite{taheri2020grabnet, karunratanakul2020grasping, karunratanakul2021skeletondriven, zhu2021toward}. \cite{jiang2021graspTTA} propose a hybrid method, where predicted contact maps on objects are used to refine an initial grasp prediction.  
Some methods have combined these two directions, for example by leveraging contact information to post-process noisy hand pose predictions \cite{grady2021contactopt}. \cite{zhang2021manipnet} generate local grasp motions, given the global motion of the hand and object. Similarly, \cite{ye2012synthesis} synthesize hand grasps given full-body and object motions. 
In summary, all of these works focus on generating static grasps and are purely data-driven. In our work, however, we take into consideration the dynamic nature of human-object interaction and consider the physical plausibility of dynamic grasp-based hand-object interactions by leveraging a physics-driven simulation. 

\textbf{Dexterous Hand Control} Different approaches have been used for controlling dexterous hands. 
Learning-based methods most often resort to an anchored hand for in-hand manipulation tasks \cite{openai2018dexterous,huang2021geometry, chen2021simple}, which removes the complexity of generating collision-free trajectories, or rely on expert demonstrations \cite{garciahernando2020iros,christen2019hri,rajeswaran2017rss, qin2021dexmv, hsiao2006imitation}, which can be costly to obtain. \cite{rajeswaran2017rss} collect expert trajectories via teleoperation, which they leverage in an RL setup to learn complex manipulation tasks. \cite{garciahernando2020iros} obtain noisy expert demonstrations from videos and use residual RL to correct the inputs for hand-object interaction tasks. In contrast, we only require a single frame grasp label per sequence. Similar to our work, \cite{christen2019hri} use a parameterized reward function from single data labels for human-robot interactions, but assume a fixed hand to interact with. \cite{jiang2021dash} propose a modular human manipulation framework, but focus on learning power-grasps for picking up objects. \cite{mandikal2020graff} intrinsically motivate a policy to grasp in the affordance region of objects. However, since the policy is only incentivized to grasp in a certain region, the fingers often end up in unnatural configurations. In their follow-up work \cite{mandikal2021dexvip}, the authors address this issue by formulating a reward based on hand-object interaction videos. However, the focus is on a single "consensus" grasp reference per object.  In our work, we propose a method that learns natural object interactions and generates a wider variety of grasps by explicitly conditioning on the desired contact points and hand pose.

\textbf{Physics-aware Inference}
Several recent works have introduced physical awareness to improve purely data-driven approaches \cite{luo2021dynreg,megzhanni2021physshapes,rempe2020eccv,shimada2021neuralcap,shimada2020tog, yuan2021simpoe, ehsani2020force}. \cite{megzhanni2021physshapes} use a physics simulation to validate the plausibility of a generative model for objects via a stability measure. \cite{ehsani2020force} learn to reason about contacts and forces in hand-object interaction videos by leveraging a physics simulation for supervision. To improve the task of human-pose reconstruction from videos, different methods have added physics-based modules to correct the output of a human-pose estimation model. This is achieved either in a post-processing optimization framework \cite{rempe2020eccv,shimada2020tog}, with an approxmation of physics \cite{shimada2021neuralcap}, or via a reinforcement learning policy that directly corrects the pose estimate \cite{yuan2021simpoe}. \cite{luo2021dynreg} regulate a data-driven policy for ego-centric pose estimation with a physics-based policy. They include full-body interactions with larger objects, such as pushing a box.
In contrast to these works, we introduce the novel task of dynamic hand-object interactions, which involves more fine-grained control of the dexterous human hand and has to adhere to the dynamics and displacement of the object of interest. The task also introduces additional complexities due to the increased amount of collision detection queries required for accurately modeling the contacts. To the best of our knowledge, ours is the first method that studies this task and constitutes an important first step into an important direction for human-object interaction. 

\section{Method}
\label{method}
We propose \emph{D-Grasp}, an RL-based approach that leverages a physics simulation for the \emph{\taskname}~task (\Fig{method_overview}). Our model requires a static grasp label consisting of the hand's 6D global pose and local pose for the fingers. We split the task into two distinct phases, namely a \emph{grasping} and a \emph{motion synthesis} phase. In the \emph{grasping} phase, the hand needs to approach an object and find a physically-plausible and stable grasp. In the \emph{motion synthesis} phase, the hand has to bring the object into the 6D target pose while the grasping policy retains a stable grasp on the object. Therefore, the grasping policy and motion synthesis module act concurrently in this phase. To this end, we follow a hierarchical framework that functionally separates the grasping from the motion synthesis. 

In the next section, we define the task setting and provide background on RL and the physics simulation. Thereafter, we present both the \textit{grasping} and \textit{motion synthesis} phases of our method in Sections \ref{method_grasp} and \ref{method_motion}, respectively.


\subsection{Task Setting}
In the \textit{\taskname} task, we are given a 6D global pose $\posesix_h$ and 3D local pose $\posevec$ of a hand, and an object pose $\posesix_o$, where the 6D poses consist of a rotation and translation component $\posesix=[\mathbf{q}|\mathbf{t}]$. Given a label of a static grasp $\grasplabel=(\overline{\mathbf{q}}_h, \overline{\posesix}_h, \overline{\posesix}_o)$, the goal is to grasp the object and move it into a 6D goal pose  $\posesix_g$. The grasp label consists of the 6D global pose of the hand $\overline{\posesix}_h$ and object  $\overline{\posesix}_o$, as well as the target hand pose $\overline{\mathbf{q}}_h$ at the instance of the static grasp. 


\paragraph{Simulation Setup}
\label{sec:simulation_stup}
To  approximate a human-like hand in the physics engine, we create a controllable hand model and integrate information obtained from a statistical parametric hand model (i.e., MANO \cite{romero2017mano}). We extract the skeleton of the hand to get the relative joint positions and add joint actuators for the control of the hand. Finally, we restrict the joints to be within reasonable limits. In our implementation, we use a unified hand model corresponding to the mean MANO shape. 
Objects are modeled via meshes from the respective datasets \cite{chao2021dexycb,hampali2020ho3d}. To further speed up the physics simulation, we approximate simple objects with primitive shapes via mesh alignment during training (e.g., a soup can is approximated by a cylinder). For more complex shapes, we use mesh decimation to reduce the number of vertices. For further details, please refer to \supmat. 

\paragraph{Reinforcement Learning}
We follow the standard formulation of a Markov Decision Process (MDP).
The MDP is defined as a tuple $\mdp = \{\StatesSet, \ActionsSet, , \mathcal{R},  \discountRate, \TransitionSet, \rho_0, \}$, where $\StatesSet$ and $\ActionsSet$ are state and action spaces, respectively. $\mathcal{R} : \StatesSet \times \ActionsSet \to \mathbb{R}$ is the reward function, $\gamma\in [0,1]$ a discount factor, $\TransitionSet : \StatesSet \times \ActionsSet \to \StatesSet$ the deterministic transition function of the environment and $\rho_0 = p(\statevec_0)$ the initial state distribution.
We aim to find a probabilistic policy $\policyvec(\actions_t|\statevec_t)$ with $\actions_t \in \ActionsSet$ and $\statevec_t \in \StatesSet$, maximizing the expected return $\mathbb{E}_{\actions_t \sim \policyvec(\cdot | \statevec_t), \statevec_0 \sim \rho_0} \left [\sum_{i=0}^{T}\discountRate^{i}\mathcal{R}(\statevec_t, \actions_t) \right ]$ with $\statevec_{t+1} = \TransitionSet(\statevec_t, \actions_t)$ at each timestep $t$.

\paragraph{State Space}
\label{method_state}

The state $\statevec = (\posevec, \dot{\mathbf{q}}_h, \forcevec, \posesix_h, \dot{\posesix}_h, \posesix_o, \dot{\posesix}_o)$ entails proprioceptive information about the hand pose in the form of joint angles $\posevec$ and joint angular velocities $\dot{\posevecgen}_h$, the forces between the hand and object $\forcevec$, the 6D pose of the wrist $\posesix_h$ and the global 6D pose of the object $\posesix_o$ with their corresponding velocities $\dot{\posesix}_h$ and $\dot{\posesix}_o$. 
States are expressed with respect to a fixed global coordinate frame. 
We show experimentally that learning from the full state space can impede learning over several different grasp labels (\secref{exp_ablation}). We therefore propose a representation that enables learning of the task in \secref{method_feat_grasp}. 

\paragraph{Action Space}
\label{method_action}
We define an action space to control the hand in the physics simulation. The fingers are controlled via one actuator per joint for a total of 45 actuators, to which we add 6 DoF to control the global pose. We employ PD-controllers that take reference joint angles $\posevecgen_{\text{ref}}$ as input and compute the torques that should be applied to the joints \jh{what did you do with the orientation?}:
\begin{align}
    \torques &= k_p (\posevecgen_{\text{ref}} - \posevecgen) + k_d \dot{\posevecgen} \\
    \posevecgen_{\text{ref}} &= \posevecgen_b + \actions.
\end{align}
The policy $\policyvec$ outputs actions $\actions$, which are residual actions that change a bias term $\posevecgen_b$. For the finger joints, the bias term is equivalent to the current joint configuration $\posevecgen_b = \posevec$. We found this formulation to lead to smoother finger motion and therefore more stable grasps compared to the policy directly predicting $\posevecgen_{\text{ref}}$. \update{Note that for simplicity's sake, we use the notation $\mathbf{q}_b$ for all joints, although the first three DoF are translational joints.}

\subsection{Physically Plausible Grasping}
\label{method_grasp}


Here we discuss the \textit{grasping} phase. The goal is to approach an object and find a physically plausible grasp. A careful design of the model's input representation is key to learning a successful model for hand-object interactions~\cite{zhang2021manipnet}, which we show in our ablations (\secref{exp_ablation}). Therefore, we introduce a feature extraction layer that converts the information from the physics simulation and grasp label into a suitable representation for model learning.

\subsubsection{Feature Extraction for Grasping}
\label{method_feat_grasp}
\begin{figure}[t]
\begin{center}
   \includegraphics[width=0.49\textwidth]{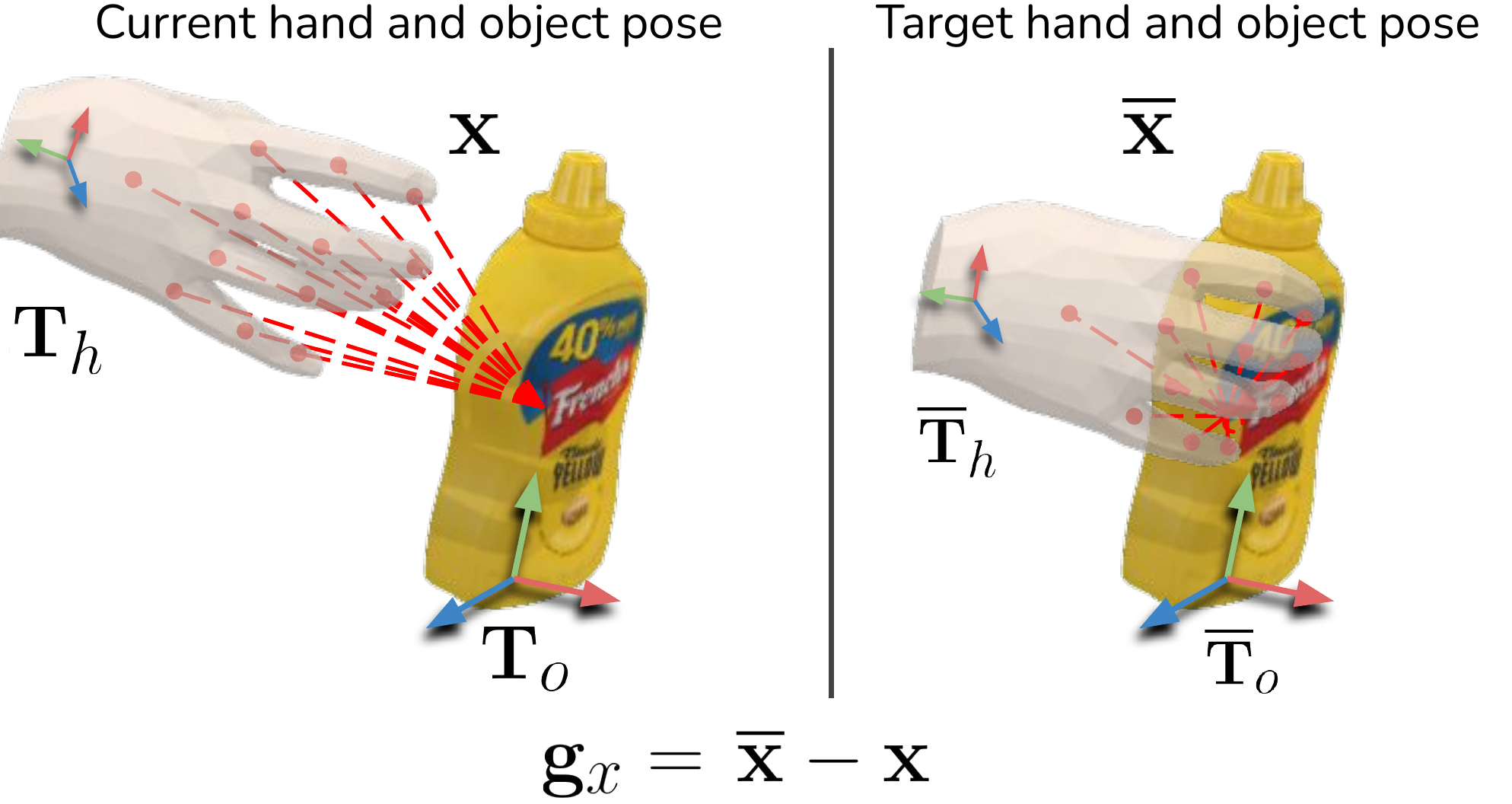}
   \vspace{-1.0cm}
\end{center}
   \caption{\textbf{Target Distance Component $\mathbf{g}_x$.} It incentivizes the policy to reach target points close to the grasp reference label $\grasplabel$. We extract the object-relative target 3D joint positions $\mathbf{\overline{x}}$ from $\grasplabel$ and compute the distance between $\mathbf{\overline{x}}$ and the current 3D joint positions $\mathbf{x}$ relative to the object's origin. We then convert  $\mathbf{g}_x$ into wrist-relative coordinates $\widetilde{\mathbf{g}}_x.$}
   \vspace{-0.0cm}
\label{fig:goal}
\end{figure}

Rather than directly conditioning the policy on the state, we apply a feature extraction layer $\phi(\statevec,\grasplabel)$ that takes the state and grasp label as input. For consistency, we can reformulate the policy as $\gpolicyvec(\actions|\phi(\statevec,\grasplabel))$ (\figref{fig:method_overview}). 
The function $\phi(\cdot)$ processes information from the grasp label, and applies coordinate frame transformations to achieve invariance w.r.t. global coordinates by transforming it to object-relative coordinates. To this end, the feature extraction layer receives the state  $\statevec = (\posevec, \dot{\mathbf{q}}_h, \forcevec, \posesix_h, \dot{\mathbf{T}}_h, \posesix_o, \dot{\posesix}_o)$ and grasp label $\grasplabel=(\overline{\mathbf{q}}_h, \overline{\posesix}_h, \overline{\posesix}_o)$ as input. Its output is defined as: 
\begin{equation}
     \phi(\statevec,\grasplabel) =(\posevec, \dot{\mathbf{q}}_h, \forcevec, \widetilde{\posesix}_h, \widetilde{\posesix}_o, \dot{\widetilde{\posesix}}_o, \dot{\widetilde{\posesix}}_h, \widetilde{\posvec}_{o}, \widetilde{\posvec}_z, \goals).
\end{equation}
The terms $\posevec$ and $\dot{\mathbf{q}}_h$ are the local joint angles and velocities, whereas $\forcevec$ represents contact force information. The remaining components are expressed in the wrist's reference frame (denoted by $\widetilde{\cdot}$ ): the object's 6D pose $\widetilde{\posesix}_o$ and its linear and angular velocities $\dot{\widetilde{\posesix}}_o$, the hand's 6D pose $\widetilde{\posesix}_h$ (relative to the initial wrist pose) and its linear and angular velocity $\dot{\widetilde{\posesix}}_h$, and the displacement of the object from its initial position $\widetilde{\posvec}_{o}$. Furthermore, $\widetilde{\posvec}_z$ introduces awareness of the vertical distance to the surface where the object rests. Lastly, we include the goal components $\goals=[\widetilde{\goalvec}_x|\widetilde{\goalvec}_q|\goalvec_c]$, which incentivize the model to reach contact points on the object. We show that these goal components are crucial for achieving stable grasps in \secref{exp_ablation}. 
More specifically, the term $\widetilde{\goalvec}_x$ measures the 3D distance between the current and the target 3D positions (\figref{fig:goal}), $\posvec$ and $\overline{\posvec}$, respectively. Here, all joints and the fingertips are in the wrist's coordinate frame. Importantly, we compute object-relative target positions from the label $\grasplabel$ in order to be invariant to the object 6D pose during the grasping phase. 


The term $\widetilde{\goalvec}_q$ represents the angular distance between the current rotations $\posevec$ and target rotations $\overline{\posevecgen}_h$ for the joints and the wrist. 
Finally, $\goalvec_c$ includes the target contact vector $\goalvec_{c}$, i.e., which finger joints should be in contact with the object. 
A more detailed description about how we extract target contacts, the applied reference frame conversions, and the coordinate representation for individual components of the state or goal space is provided in \supmat.

\subsubsection{Reward Function for Grasping}
\label{method_reward}
To incentivize the policy to learn the desired behavior, we need to define a reward function. In our method, we formulate it as follows:
\begin{equation}
    r = w_{x} r_{x} +  w_{q} r_{q} +  w_{c} r_{c} +  w_{\text{reg}} r_{\text{reg}}.
\end{equation}
It comprises a combination between position, angle, contact and regularization terms, respectively. We weigh the reward components with the factors $w_{x},  w_{q},  w_{c},  w_{\text{reg}}$.\\
The position reward $r_{x}$ measures the weighted sum of distances between the target $\overline{\mathbf{x}}$ and the current 3D positions $\posvec$ for every joint (including the wrist):
\begin{equation}
    r_{x} = \sum^{J}_{j = 1} w_{x,j}\| \overline{\posvec}_j - \posvec_j\|^2.
\end{equation}
Similarly, the pose reward $r_q$ measures the distance between the current pose and the corresponding target pose in Euler angles and corresponds to the L2-norm of the feature $\widetilde{\goalvec}_q$:
\begin{equation}
   r_{q} = \|\widetilde{\goalvec}_q\|,
\end{equation}
The contact reward $r_{c}$ is extracted from the finger parts that should be in contact with the object. Specifically, it is computed as the sum of two terms. The first one represents the fraction of target contacts that the agent has achieved. The second term rewards the amount of force exerted on desired contact points, capped by a factor proportional to the object's weight $m_{o}$ through a factor $\lambda$:

\begin{equation}
    r_{c} = \frac{\widetilde{\goalvec}_c^\top \mathbf{I}_{\forcevec>0}}{\widetilde{\goalvec}_c^\top \widetilde{\goalvec}_c} + \min(\widetilde{\goalvec}_c^\top \forcevec, \lambda m_{o}).
\end{equation}
Finally, the reward $r^{\text{reg}}$ involves regularization terms on the hand's and object's linear and angular velocities:
\begin{equation}
r_{\text{reg}} =  w_{\text{reg},h} \| \dot{\widetilde{\posesix}}_h \|^2
+  w_{\text{reg},o} \| \dot{\widetilde{\posesix}}_o \|^2.
\end{equation} 

\begin{table*}[t]
	\centering
	\resizebox{0.8\textwidth}{!}{%
		\begin{tabular}{l|ll|ccc|ccc}
		
			\toprule
            & & \multirow{2}{*}{Models} & \multicolumn{3}{c}{Training set} & \multicolumn{3}{c}{Test set} \\
			& &  & \scriptsize{Success$\uparrow$} &  \scriptsize{SimDist [mm/s]$\downarrow$} &  \scriptsize{Interp. [$cm^3$]$\downarrow$} &  \scriptsize{Success$\uparrow$} &  \scriptsize{SimDist [mm/s]$\downarrow$} &  \scriptsize{Interp. [$cm^3$]$\downarrow$}  \\
			
			\midrule
            \parbox[t]{2mm}{\multirow{5}{*}{\rotatebox[origin=c]{90}{DexYCB}}}
			& \parbox[t]{2mm}{\multirow{3}{*}{\rotatebox[origin=c]{90}{\scriptsize{MC}}}}  
			& GT+PD & 0.31 & $13.4 \pm 9.2$ & 4.59 & 0.35 & 13.1 $\pm 9.1$ & 4.41 \\
            & & \greycellcol GT+IK & \greycellcol 0.39 & \greycellcol $11.8 \pm 9.4$ & \greycellcol 9.23 & \greycellcol 0.50 & \greycellcol $9.1 \pm 8.5 $ & \greycellcol 9.74\\
            
            & & \textbf{Ours} & \textbf{0.70}  & $\mathbf{5.8 \pm 7.4}$  & \textbf{1.75} & \textbf{0.63}  & $\mathbf{8.0 \pm 8.1}$  & \textbf{1.77}\\
            
            


			\cmidrule{2-9}
			
		    & \parbox[t]{2mm}{\multirow{2}{*}{\rotatebox[origin=c]{90}{\scriptsize{SYN}}}} 
			& Jiang \textit{et. al} \cite{jiang2021graspTTA}+PD & 0.25 & $12.4 \pm 6.4$ &  4.92 & 0.24 & $12.7 \pm 6.5 $ &  4.94 \\
			
			& & \greycellcol \textbf{Ours} & \greycellcol \textbf{0.75} & \greycellcol $\mathbf{3.9 \pm 7.2}$ & \greycellcol \textbf{2.84} &  \greycellcol \textbf{0.73} & \greycellcol $\mathbf{4.6 \pm 6.7}$ & \greycellcol\textbf{2.81} \\

			\midrule
			\parbox[t]{2mm}{\multirow{4}{*}{\rotatebox[origin=c]{90}{HO3D}}}
		    & \parbox[t]{2mm}{\multirow{2}{*}{\rotatebox[origin=c]{90}{\scriptsize{SYN}}}} 
			& Jiang \textit{et. al} \cite{jiang2021graspTTA}+PD & 0.31 & $ 10.0 \pm 6.6$ & 5.21 & 0.30 & $10.6 \pm 6.8 $ & 5.40\\
			& & \greycellcol \textbf{Ours} & \greycellcol \textbf{0.73} & \greycellcol $\mathbf{4.4 \pm 7.4}$ & \greycellcol \textbf{3.33} & \greycellcol \textbf{0.71} & \greycellcol $\mathbf{4.9 \pm 6.6}$ & \greycellcol \textbf{3.40} \\
			
			\cmidrule{2-9}
			
		    & \parbox[t]{2mm}{\multirow{2}{*}{\rotatebox[origin=c]{90}{\scriptsize{IMG}}}} 
			& Grady \textit{et. al} \cite{grady2021contactopt}+PD & 0.67 & $ 5.1 \pm 6.1$ & 14.94 & 0.60 & $6.5 \pm 5.8$ & 14.00 \\
			& & \greycellcol \textbf{Ours} & \greycellcol \textbf{0.88} & \greycellcol $\mathbf{1.4 \pm 3.4}$ & \greycellcol \textbf{2.67} & \greycellcol \textbf{0.81}  & \greycellcol $\mathbf{1.9 \pm 3.6}$ & \greycellcol \textbf{2.08} \\
			
			\bottomrule
			
		\end{tabular}%
	}
	\vspace{-0.1cm}
	\caption{\textbf{Static grasp evaluation}. 
	We compare our model with grasp samples \camera{from the DexYCB dataset (MC), generated samples by a grasp synthesis method on the DexYCB and HO3D object sets (SYN), and samples extracted from an image-based hand pose estimator (IMG)}. We evaluate the baseline grasps in the simulation via PD-control (*+PD) directly or after de-noising via inverse kinematics (*+IK) for the motion capture data. We observe that our method outperforms the baselines in all metrics and conditions. The results indicate that static grasp references 1) will not lead to stable grasps when evaluated in a physics simulation and 2) suffer from interpenetration. Our method improves the interpenetration and learn stable grasps in a dynamic setting.
	}
	\label{tab:exp1_train}
\end{table*}{}

\subsubsection{Wrist-Guidance Technique}
\label{wrist-guidance}
To control the global pose during the grasping phase, we introduce a simple but effective technique which we call \textit{wrist-guidance}. \update{Intuitively, we bias the hand to approach the object. To achieve this, we leverage the object-relative target pose, of the hand on the object, obtained from the grasp label $\grasplabel$. We then use it as a bias term in the PD-controller of the global 3DoF position. In other words, we set the bias term of the first 3DoF (the translational joints) to $\posevecgen_b=\overline{\mathbf{x}}_h$ (\secref{method_action}), where $\overline{\mathbf{x}}_h$ is the target position which we extract from the label. 
We find that this technique leads to better performance and faster convergence than using the previous joint positions as bias (\secref{method_action}), which we show in ablations in \secref{exp_ablation}.} 
\SC{I'm a bit unsure here about the depiction of the variables, we are just using the first three dimensions of $q_b$ and $x_h$, is it enough to indicate it in the text or do we need additional indexing?}\oh{I think its fine as is. You could also hand-wavily say: $\posevecgen_b\subset\overline{\mathbf{x}}_h$ without specifiying exactly which components you use.}

\subsection{Motion Synthesis}
\label{method_motion}
We now introduce the motion synthesis module, which is responsible for moving the object from an initial 6D pose into a target 6D pose. It controls only the movement of the wrist, i.e., the first 6DoF of the controllable hand model. In this phase, both the grasping policy described in \secref{method_grasp} and the motion synthesis module are executed concurrently. While the grasping policy maintains a stable grasp, the motion synthesis module takes over the control of the 6D pose of the hand.  Similar to the grasping policy, we propose a feature extraction layer that incentivizes the model to move the hand to a target pose with the object in-hand.  

\camera{To control the global hand motion, we estimate a 6D target pose for the hand: $\widehat{\mathbf{T}}_h = \psi(\statevec, \posesix_g, \mathbf{D})$.
In particular, we estimate the global target hand pose $\widehat{\mathbf{T}}_h$ \update{by computing the distance between the object's current 6D pose $\posesix_o$ and the target 6D pose $\overline{\posesix}_o$. We then translate and rotate the hand according to the displacement using closed-loop control. Hence, the displacement is recomputed after every action}. For more details, please refer to \supmat.}

\section{Experiments}
We conduct several experiments to analyse the performance of our method. We first introduce the data and experimental details in Sections \ref{sec:exp_data} and \ref{sec:exp_setup}. Next, we show that our method can learn stable grasps and correct imperfect labels in \secref{exp_grasp}. Lastly, we evaluate the motion synthesis task and provide ablations to highlight the importance of our method's components in Sections \ref{exp_motion} and \ref{exp_ablation}.
\subsection{Data}
\label{sec:exp_data}

\paragraph{DexYCB} We make use of the DexYCB dataset \cite{chao2021dexycb}. The dataset consists of 1000 sequences of object grasping, with 10 different subjects and 20 YCB objects \cite{calli2015ycb}. We filter out all left handed sequences and create a random 75\%/25\% train/test-split over all sequences and subjects.
The data sequence contains 6D global poses for the hand and objects in the camera frame and the local joint angles, hence providing sequences of 
$\begin{Bmatrix} (\overline{\mathbf{q}}_h, \overline{\posesix}_h, \overline{\posesix}_o)
\end{Bmatrix}_{t=1}^T$. The data also includes meshes for the hand and objects, and the camera parameters.
We determine the grasp label based on the object's displacement with regards to its initial position. The time-step with an object displacement greater than a pre-determined threshold is chosen to be the target grasp $\grasplabel$. Furthermore, we use a recent state-of-the-art grasp synthesis method \cite{jiang2021graspTTA} to generate grasp labels for all the objects in DexYCB and create a 400/200 label train/test-split.

\paragraph{HO3D}
We use generated grasp labels from static grasp synthesis \cite{jiang2021graspTTA} or \camera{from an image-based pose estimator after offline optimization \cite{grady2021contactopt} for the  HO3D objects}. We create a train/test-split that is proportional to the DexYCB split, which results in a 200/100 label train/test-split.

\subsection{Experimental Details}
\label{sec:exp_setup}
We train policies by using our implementation of the PPO algorithm \cite{schulman2017proximal} and run simulations in \raisim \cite{hwangbo2018raisim}. 
For each sequence, we initialize the environment with an object and a grasp label. The hand is initialized with a pose from earlier steps at a pre-determined distance from the object. \camera{First, we train the grasping policy with all training labels and objects}. Then we continue with the motion synthesis component given the pretrained grasping policy. \\
We evaluate physical plausibility of a grasp in terms of stability and interpenetration on a set of unseen grasp labels \update{and unseen objects}. We define a set of complementary metrics to quantify performance extensively.


\subsubsection{Metrics}
\label{exp_metrics}
\textbf{Success Rate:} We define the success rate as the primary measure of physical plausibility. 
It is measured as the percentage of sequences which maintain a stable grasp, i.e., where the object does not slip for a period of time. \\
\textbf{Interpenetration:} We calculate the amount of hand volume that penetrates the object. \camera{We compute it using the vertices of the MANO mesh \cite{romero2017mano} and the high-resolution object meshes.} \\ 
\textbf{Simulated Distance:} Similar to the metric proposed in \cite{jiang2021graspTTA}, we compute the mean displacement of the object. Instead of measuring the absolute displacement, we report the mean displacement in mm per second. \\
\textbf{Contact Ratio:} For the ablation study, we measure the ratio between the target contacts defined via the grasp label $\grasplabel$ and the contacts achieved in the physics simulation. \\
\textbf{MPE:} The mean position error between the object's position and target 3D position (for motion synthesis).\\
\textbf{Geodesic:} The angular distance between the object's \update{current} and target orientation (for motion synthesis).

\begin{figure}[t]
\begin{center}
   \includegraphics[width=0.5\textwidth]{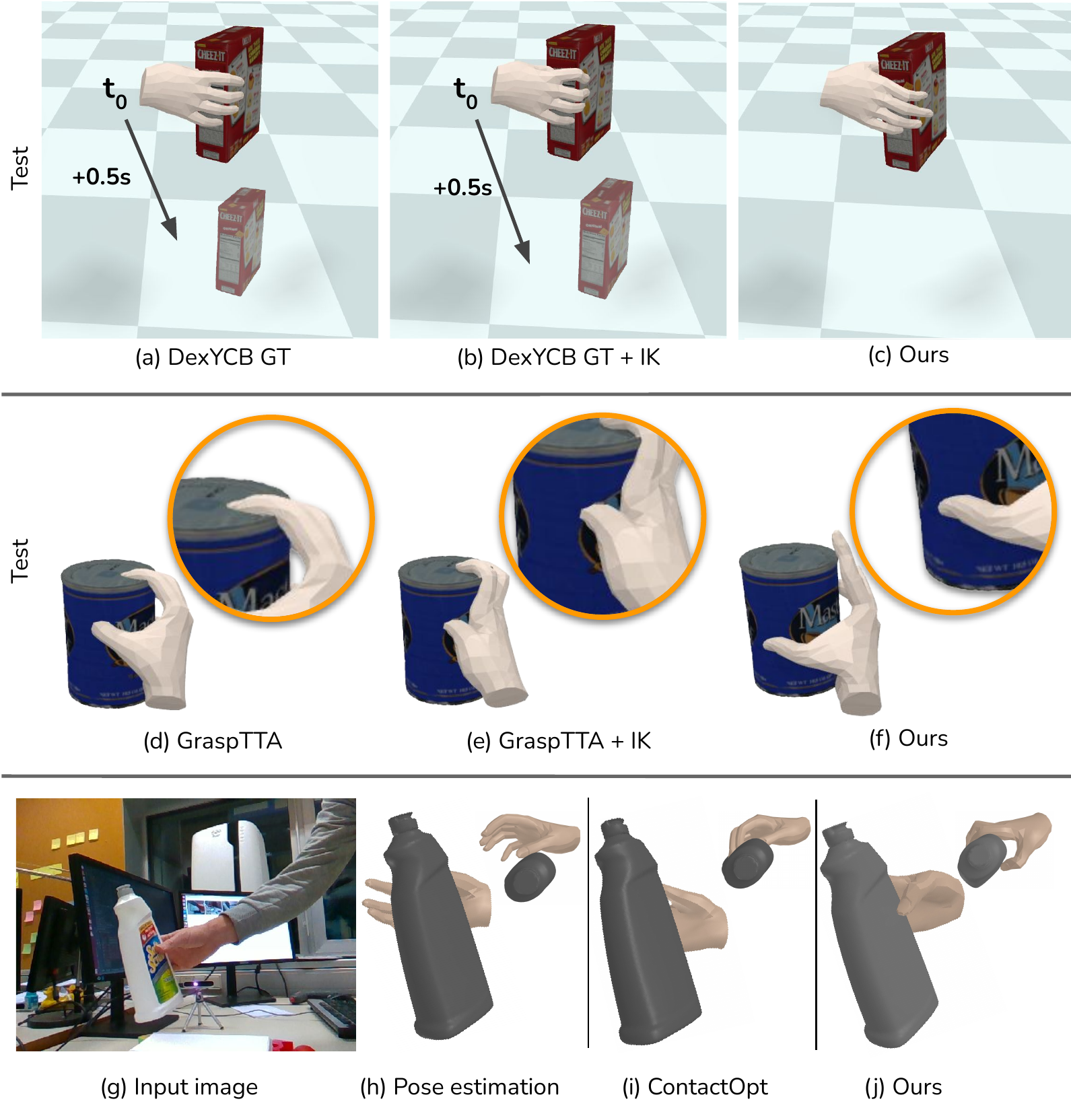}
  \vspace{-0.8cm}
\end{center}
   \caption{\textbf{Qualitative evaluation}. (a)-(c): static grasp labels often do not lead to stable grasps when evaluated in a physics simulation (a-b), which can be successfully corrected by our method (c). For an animated demonstration, please see the \textbf{video} in \supmat. (d)-(f): showcases artifacts such as interpenetration when using a state-of-the-art grasp synthesis method \cite{jiang2021graspTTA} (d-e). Our method (f) can correct such cases and generate physically-plausible grasps. \camera{(g)-(j): using images (g) to estimate an initial grasp (h). Physically implausible poses occur even with corrections via offline optimization (i), which can be corrected by our method (j).}}
\label{fig:grasp_qual}
\end{figure}

\subsubsection{Baselines}
\label{exp_baselines}
\textbf{*+PD:} Similar to \cite{jiang2021graspTTA}, we place the object into the hand via the grasp label. We then attempt to maintain the grasp using PD-control in the physics simulation.\\
\textbf{*+IK:} We employ an offline optimization to correct for imperfections (i.e., minor distances or penetrations) in the label. The improved samples are passed to the PD-control. \\
\textbf{Flat-RL:} We employ an RL baseline that does not separate the grasping from the motion synthesis phase, but trains the full dynamic grasp synthesis task end-to-end.\\
\textbf{Ours+static grasp:} In this variant, we use our grasping policy for the grasping phase. During motion synthesis, we use PD-control to maintain the pose while the grasping policy is frozen and not actively interacting with the object.

\begin{table}[t]
	\centering
	\resizebox{\columnwidth}{!}{%
		\begin{tabular}{l|ccc}
		
			\toprule

			Models & Success $\uparrow$ & SimDist [mm/s] $\downarrow$ & Interpenetration [$cm^3$]$\downarrow$  \\
			
			\midrule
            GT+PD & 0.30 & $13.7 \pm 9.2$ & 4.41 \\
            \rowcolor{Gray}
            GT+IK & 0.38 & $11.7 \pm 9.4$ & 9.08 \\
            \textbf{Ours} & \textbf{0.56} & $\mathbf{9.0 \pm 10.4}$ & \textbf{1.74}\\
			
			\bottomrule
		\end{tabular}%
	}
    \vspace{-0.1cm}
	\caption{\camera{\textbf{Generalization}. We evaluate generalization to unseen objects and compare our model with the baselines. We create six different test sets of three objects each, which we leave out during training. We report the average performance over all test sets. 
	}}
	\label{tab:exp1_gen}
\end{table}{}

\subsection{Grasping Objects}
\label{exp_grasp}
In this experiment, we show that our method can learn to achieve stable grasps and that static grasp reference data is inherently bound to fail in a dynamic setting. We first train with labels from DexYCB \cite{chao2021dexycb} and further demonstrate that our approach also works with, and improves upon, labels obtained from state-of-the-art grasp synthesis method \cite{jiang2021graspTTA}, on both the DexYCB and HO3D object sets. \camera{Lastly, we present results using an image-based hand pose estimator on HO3D images and labels from ContactOpt \cite{grady2021contactopt}}. 

We present quantitative evaluations in \tabref{tab:exp1_train} and qualitative results in \figref{fig:grasp_qual}. Compared to the baselines, our method is able to achieve significantly better performance on all the metrics. Importantly, the grasping policy can improve the success rate, while minimizing interpenetration (an important metric in the grasp synthesis literature).  \camera{We note that our method achieves 0 interpenetration loss when evaluated in the physics simulation. In \tabref{tab:exp1_train}, however, we report interpenetration on the original MANO hand model and detailed object meshes.
For computational efficiency during training, the hand model and the object meshes are simplified in the physics simulation (\secref{sec:simulation_stup}), limiting the performance of our model when evaluated in the original setting with regards to interpenetration.} \camera{We found no improvement with IK for the generated (SYN) or image-based (IMG) experiments and hence omit it from the results. The improved performance in the image setup compared to other settings is due to the high-quality grasp references from \cite{grady2021contactopt}, which already optimizes for contact}. In general, there is a performance drop when moving to unseen test labels. We also find that our approach may struggle with thin objects which are difficult to grasp on a surface. For a detailed analysis and failure cases, we refer to \supmat.

\camera{\paragraph{Generalization to Unseen Objects}
To evaluate the generalization performance on unseen objects, we train and test our model on six separate train/test splits with varying complexity. Each test set consists of three objects from the DexYCB dataset. The remaining objects are used for training a policy. We average the results over all test sets and report the results in \tabref{tab:exp1_gen}. While there is room for improvement in overall success rate, our method outperforms the baseline in all metrics. We provide a more detailed analysis in \supmat.}

\subsection{Motion Synthesis}
\label{exp_motion}
We now demonstrate our method's ability to synthesize motions with the grasped object in hand. The goal of this task is to grasp an object and generate a trajectory that brings the object to a target 6D pose. 
We use a subset of representative YCB objects and create a test set with 100 randomly sampled, out-of-distribution poses $\mathbf{T}_g$.
We compare against a standard RL baseline (Flat-RL) and a variant of our method that only \emph{maintains} the pose instead of actively grasping the object (Ours+static grasp). We also compare against a learning-based motion synthesis policy (Ours+learned policy).
As shown in \tabref{tab:motion}, the hierarchical separation in our method is crucial for solving the task. Moreover, the decrease in performance when the hand pose is simply maintained (Ours+static grasp) solidifies the contribution of our approach. This implies that active control of the hand throughout the sequence is mandatory to maintain a stable grasp. Lastly, our method outperforms the learning-based variant (Ours+learned policy) of our motion synthesis module by a large margin on both metrics. \



\begin{table}[t]
	\centering
	\resizebox{0.8\columnwidth}{!}{%
		\begin{tabular}{ll|cc}
		
			\toprule

			& Models & MPE [mm] $\downarrow$ & Geodesic [rad.] $\downarrow$  \\
			
			\midrule
            
            
            & Flat-RL & 0.55 & 1.66 \\
            & \greycellcol Ours+static grasp & \greycellcol 0.45 & \greycellcol 1.46 \\
            & Ours+learned policy & 0.30 & 0.92 \\
            & \greycellcol \textbf{Ours} & \greycellcol \textbf{0.08}  & \greycellcol \textbf{0.52}  \\
			
			\bottomrule
		\end{tabular}%
	}

	\caption{\textbf{Evaluation of motion synthesis}. We compare our model with a standard RL baseline (Flat-RL) and different variants of our method. We observe that our hierarchical framework outperforms Flat-RL. Furthermore, an active grasping policy during motion synthesis is key to solving the task, as indicated by the performance drop for Ours+static grasp.}
	\label{tab:motion}
\end{table}{}

\subsection{Ablations}
\label{exp_ablation}
In this experiment, we analyze different components of our method and show that they are crucial for achieving stable grasps. To this end, we ablate our method with different feature spaces and reward functions. We select a subset of representative objects and evaluate on our train-split of DexYCB  (\secref{sec:exp_data}).
To validate our feature extraction layer and in particular the goal space (\secref{method_feat_grasp}), we compare to a variant of our approach using the original state space (w/o FeatLayer) and a variant without the goal space (w/o GoalSpace). Furthermore, we evaluate our method without the contact reward (w/o ContactRew) and without the proposed wrist-guidance (w/o WristGuidance) as proposed in \secref{wrist-guidance}. \Tab{ablation} shows that each component yields considerable performance improvement. We emphasize that the contact reward and a suitable feature representation are key for achieving stable grasps.

\begin{table}[t]
	\centering
	\resizebox{\columnwidth}{!}{%
		\begin{tabular}{l|ccc}
		
			\toprule

			Models & Success $\uparrow$ & SimDist [mm/s] $\downarrow$ & Contact Ratio $\uparrow$  \\
			
			\midrule
            w/o ContactRew & 0.0 & $24.18 \pm 1.58$ & 0.02\\
            \rowcolor{Gray}
            w/o GoalSpace & 0.28 & $14.21 \pm 10.50$ & 0.18 \\
            w/o FeatLayer & 0.47 & $9.69 \pm 10.26$ & 0.21 \\
            \rowcolor{Gray}
			w/o WristGuidance & 0.58  & $7.88 \pm 10.57$ & 0.28 \\
            
            \textbf{Ours} & \textbf{0.89} & $\mathbf{4.83 \pm 1.71}$ & \textbf{0.43}\\
			
			\bottomrule
		\end{tabular}%
	}

	\caption{\textbf{Ablations}. We ablate our proposed components. All components together comprises our method. We observe that each component increases the performance significantly in all metrics.}
	\label{tab:ablation}
\end{table}{}

\section{Discussion and Conclusion}
\camera{
In this work we have made several contributions. First, we have introduced the task of \textit{dynamic grasp synthesis} for human-object interactions. To take a meaningful step into this direction, we leverage a physics simulation to generate sequences of hand-object interactions that are natural and physically plausible. We propose an RL-based solution that learns from a single external grasp label. We demonstrate that our method can learn stable grasps and generate motions with the object-in hand without slipping. Furthermore, we have provided evidence that our method can achieve generalization to unseen objects. While this proof of concept experiment indicates that our method works if a static hand pose reference for the unseen object is available, the method could be scaled to even larger train/test sets in the future. Finally, dynamics components such as friction of surfaces, inertia, or the center of mass are assumed to be known a priori, which is often not the case in real world settings. Adding a perceptual component to estimate these properties is a promising direction for future work.
}

\paragraph*{Acknowledgements:}

\begin{wrapfigure}{R}{0.35\columnwidth}
\vspace{-0.3cm}
\centering
\hspace{-.5in}
\vspace{-.3in}
\includegraphics[width=0.35\columnwidth, trim=20 100 180 180]{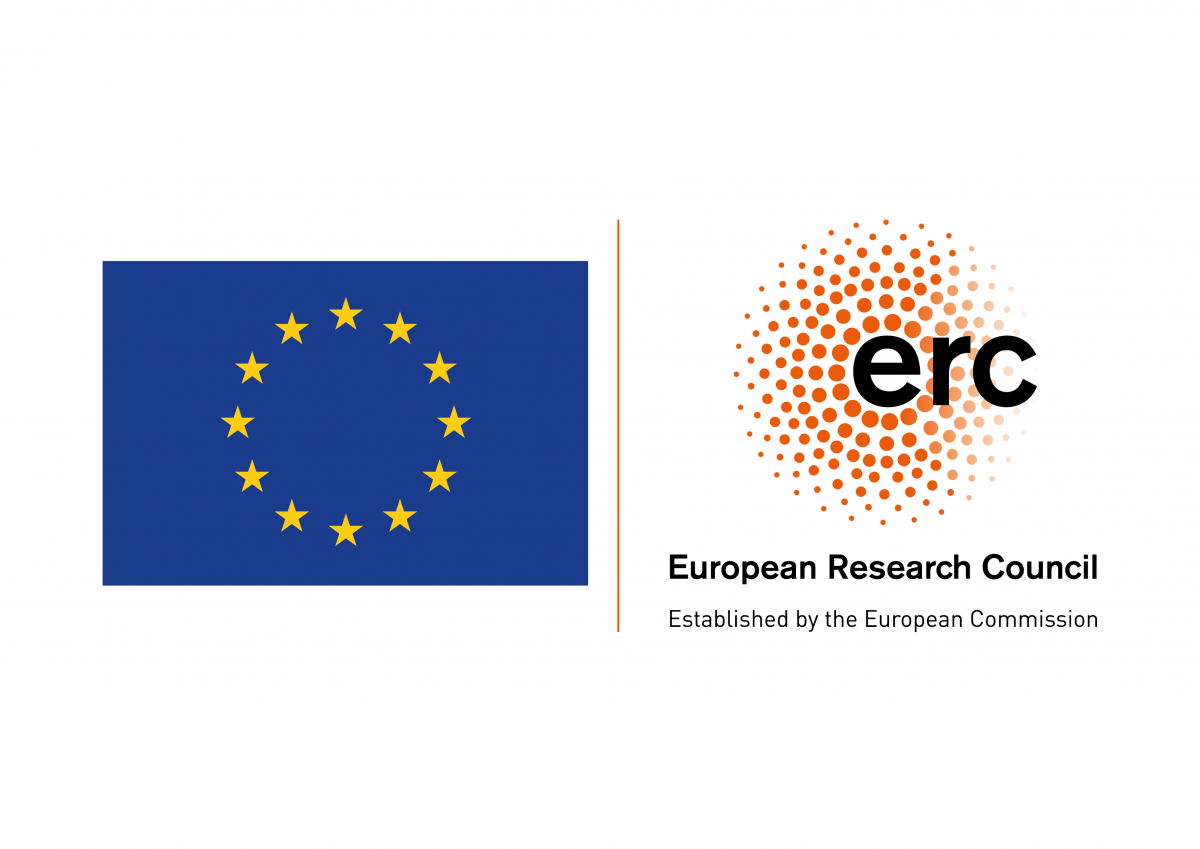}

\end{wrapfigure}
\small

This project has received funding from the European Research Council (ERC) under the European Union's Horizon 2020 research and innovation programme grant agreement No 717054. Muhammed Kocabas is supported by the Max Planck ETH Center for Learning Systems. We thank Marco Bagatella, Manuel Kaufmann, Thomas Langerak, and Adrian Spurr for the fruitful discussions and help throughout this project. Lastly, we thank Alexis E. Block for the voice-over.

\normalsize

{\small
\bibliographystyle{ieee_fullname}
\bibliography{egbib}
}


\appendix
\twocolumn[
\centering
\Large
\textbf{D-Grasp: Physically Plausible Dynamic Grasp Synthesis\\ for Hand-Object Interactions} \\
\vspace{0.5em}Supplementary Material \\

\vspace{1.0em}
] 

  

The supplementary material of this paper includes \textbf{a video} and this document. We provide more detailed descriptions of our method in \secref{app_method} and implementation details (physics simulation, baselines, metrics, and the learning algorithm) in \secref{app_impl}. Furthermore, we present additional qualitative results, as well as more detailed quantitative results in \secref{app_results}. Lastly, we discuss potential societal impacts in \secref{app_lim} and provide a glossary for the notations used in this paper in \secref{app_glossary}.  

\appendix

\section{Method Details}
\label{app_method}

We presented our method in \secref{method}. Importantly, we functionally separate the 6DoF global motion synthesis module from the grasping policy. We achieve this by explicitly separating the information flow in the feature extraction layers $\phi(\cdot)$ and $\psi(\cdot)$, similar to \cite{christen2021hide}.  We show in \secref{exp_motion} that this enables solving the complex \emph{\taskname}~task. We now provide more details on the feature extraction layers.

\subsection{Grasping Feature Extraction Details}
\label{app_feat_grasp}

We detail our method's grasping policy in \secref{method_grasp}. In this section, we provide additional details on how we extract the features of the goal space presented in \secref{method_feat_grasp}. Hence, we need to extract object-relative features from the label $\grasplabel$ in order to be invariant to the object 6D pose during the grasping phase. Since collisions with the object occur when learning a grasp, it is crucial to have a representation that is flexible with respect to the object's pose, even when its position changes. We therefore focus on explaining the goal components $\goals=[\widetilde{\goalvec}_x|\widetilde{\goalvec}_q|\goalvec_c]$.
\paragraph{Relative target positions:}
The term $\widetilde{\goalvec}_x$ measures the 3D distances between the hand's current and the target joint 3D positions $\posvec_h$ and $\overline{\posvec}_h$, respectively. Hence, to get the 3D target positions $\overline{\posvec}_h$, we utilize the label's information about the (global) 6D poses of the object $\overline{\posesix}_o$ and the hand $\overline{\posesix}_h$, as well as the target joint configuration $\overline{\mathbf{q}}_h$. Specifically, we use forward kinematics to compute the global target pose of the hand, which we then convert into the object-relative coordinate frame using $\overline{\posesix}_o$. This provides us with the 3D target positions $\overline{\mathbf{x}}_h$ for all the joints. We then apply the same procedure to the current state of the environment, using the object's current 6D pose $\posesix_o$, the hand's current 6D pose $\posesix_h$ and the hand's current joint configuration $\posevec$. This gives us the 3D joint positions of the current hand configuration $\posvec_h$ in the object-relative frame. Next, we measure the distance between the current and target joint positions:

\begin{equation*}
   \goalvec_x = \overline{\mathbf{x}}_h-\mathbf{x}_h.
\end{equation*}
Our final step consists of transforming $\goalvec_x$ into wrist-relative coordinates, finally providing us with $\widetilde{\goalvec}_x$.

\paragraph{Relative target rotations:}
The term $\widetilde{\goalvec}_q$ represents the angular distances between the current and target rotations for the joints and the wrist. For the local joint rotations, we can directly compute the distance between the current joint rotations $\posevec$ and the target joint rotations $\overline{\mathbf{q}}_h$. For the orientation of the wrist, we follow the abovementioned procedure to achieve invariance to the object pose. Hence, we convert the global 6D hand target pose $\overline{\posesix}_h$ into an object-relative target pose using $\overline{\posesix}_o$. We apply the same conversion to the current 6D hand pose $\posesix_h$ using the object's current 6D pose $\posesix_o$. We then compute the angular distance between the current and target object-relative poses. Finally, we transform the computed distance into wrist-relative frame for consistency.  \\

\paragraph{Target contacts:}
The contact goal vector $\goalvec_c=(\overline{\goalvec}_c, \mathbb{I}_{\forcevec, 
\overline{\goalvec}_{c}>0})$ is the concatenation of two vectors, namely the desired contacts $\overline{\goalvec}_{c}$ and the term $\mathbb{I}_{\overline{\goalvec}_{c}>0}$. 
To get the desired contacts for each hand joint from the grasp label, we measure the distance between all of a joint's vertices of the created meshes (\secref{sec:simulation_stup})  and all the vertices of the object mesh, which can be computed from the grasp label $\grasplabel$. Hence, for each joint $j$, the desired contacts are then determined as follows: 
\begin{equation}
    \goalvec_{c,j} =  \mathbb{I} \left [ \sum^{I}_{i = 1} \sum^{O}_{o = 1} \mathbb{I} [\|\overline{\vertexvec}_{\text{i}}-\overline{\vertexvec}_{o}\|^2<\epsilon] > 0\right].
\end{equation} 
If the distance between any vertex $\overline{\vertexvec}_{\text{i}}$ of a joint $j$ and an object vertex $\overline{\vertexvec}_{o}$ is below a small threshold $\epsilon$ (in our case 0.015m), we determine that the finger part should be in contact and hence the contact label should be equal to 1, otherwise 0. \\ 
The component $\mathbb{I}_{\overline{\goalvec}_{c}>0}$ is a one-hot encoding vector indicating which of the desired contacts $\overline{\goalvec}_{c}$ are active. Please note the redundancy in $\goalvec_c$, which may be further improved in future work.

\subsection{Motion Synthesis}

\paragraph{Ours} As described in \secref{method_motion}, we use a closed-loop control scheme to move the hand from its current 6D pose $\posesix_h$ to the estimated hand pose $\widehat{\posesix}_h$. In particular, we compute the distance between the current and estimated target 6D object pose $\Delta \widehat{\posesix}_h = (\widehat{\posesix}_o-\posesix_o)$. This term is then added to the current 6D pose of the hand and weighted by a factor $\beta$:
\begin{equation}
   \posesix_\text{pd} = \posesix_h + \beta \Delta \widehat{\posesix}_h.
\end{equation}
The term $\posesix_\text{pd}$ is then sent to the PD-controller of the simulation. The output of the PD-controller are torques that generate a motion to guide the hand to the estimated target pose $\widehat{\posesix}_h$ by recomputing $\Delta \widehat{\posesix}_h$ after each simulation update. Note that in the \emph{motion synthesis} phase, this module replaces the control of the first 6DoF of the grasping policy.

\paragraph{Ours+Learned Policy}
\label{app:motion_synthesis_feat}
For the learned variant of the motion synthesis module, we propose a feature layer $\psi(\statevec, \posesix_g, \grasplabel)$ and a motion policy $\mpolicyvec(\actions_m | \psi(\statevec, \posesix_g, \grasplabel))$. Intuitively, it is not necessary for the motion policy to know about the proprioceptive information of the hand, such as joint angles and angular velocities. Therefore, we only extract features which are relevant to the global control of the 6D hand pose $\posesix_h$. The feature extraction layer $\psi(\statevec, \posesix_g, \grasplabel)$ receives the state $\statevec$ and the 6D target pose $\posesix_g$ of the object. The output of this layer is the following:
\begin{equation}
    \psi(\statevec, \posesix_g, \grasplabel) \overline{=}({\posesix}_h, \dot{{\posesix}}_h, {\posesix}_o, \dot{{\posesix}}_o, {\goalvec}_{o,x}, {\goalvec}_{o,q}),
\end{equation}
where the first four terms include information about the 6D poses and respective velocities of the hand and object. Crucially, the features ${\goalvec}_{o,x}$ and ${\goalvec}_{o,q}$ entail information about the object's current and target pose. The term ${\goalvec}_{o,x}$ is the Euclidean distance between the object's current and target position ${\goalvec}_{o,x}={\posesix}_{o,x}-{\posesix}_{g,x}$ in global coordinates. Similarly, ${\goalvec}_{o,q}$ computes the angular distance between the object's current and target pose ${\goalvec}_{o,q}={\posesix}_{o,q}-{\posesix}_{g,q}$. For motion synthesis, we use the following reward function:
\begin{equation}
    r_m = \alpha_{x} r_{m,x} +  \alpha_{q} r_{m,q}.
    \label{rew_motion}
\end{equation}
The position reward $r_{m,x}=e_{\text{mpe}}$ measures the distance between the current and target object position (Eq. \ref{eq_mpe}). The angular reward is the geodesic distance between the object's current and target orientation $r_{m,q}=e_{\text{geo}}$ (Eq. \ref{eq_geo}). We weigh the two components with factors $\alpha_x$ and $\alpha_q$.\\
In general, we propose a learning based variant because we believe it could come in as a viable solution when the control of the global hand pose becomes more complex. In the current work, we directly control the 6D pose of the hand. In such a setting, an IK-based solution is expected to outperform a learning-based variant. In the future, one could extend our method to include a biomechanical model of a full arm. This would add inherent constraints to the hand movements and hence increase the complexity of controlling the hand successfully. On the upside, this may lead to more natural movements during the \emph{motion synthesis} phase. Hence, in such a setting a learning-based variant may outperform an IK-based solution.


\section{Implementation Details}
\label{app_impl}

\subsection{Physics Simulation}

\begin{figure}[t]
\begin{center}
   \includegraphics[width=0.7\columnwidth]{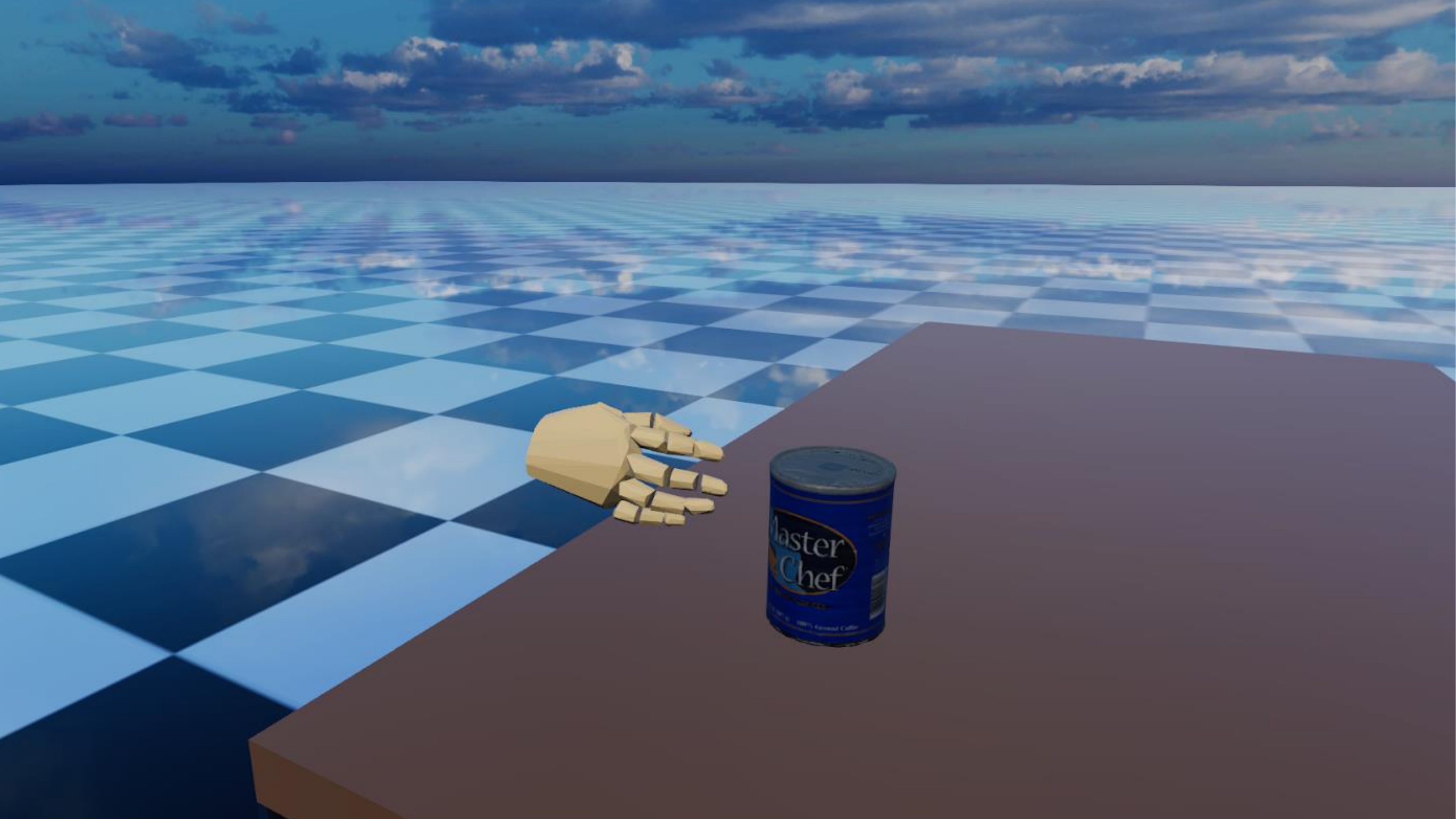}
\end{center}
   \caption{\textbf{Physics Simulation.} We create a controllable hand model and deploy it in the RaiSim physics-engine \cite{hwangbo2018raisim} to provide us with information about contacts and dynamics.}
   \vspace{-0.4cm}
\label{fig:raisim}
\end{figure}

\begin{figure}[t]
\begin{center}
   \includegraphics[width=0.5\columnwidth]{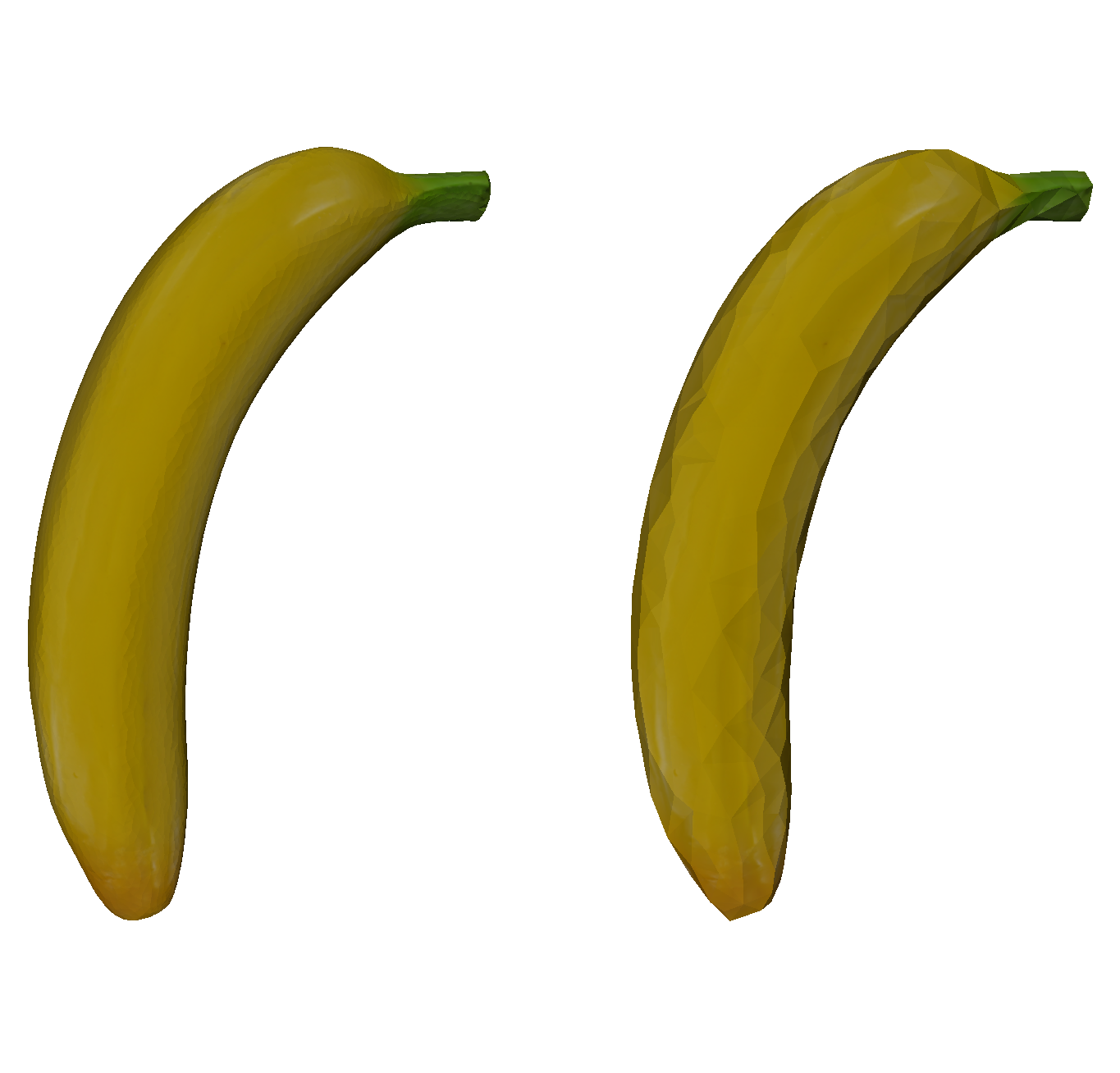}
\end{center}
\vspace{-0.5cm}
   \caption{\textbf{Mesh Decimation.} We use mesh decimation to reduce the number of vertices of the object mesh. On the left is the original object mesh, on the right the decimated mesh. This helps to speed up the physics simulation during training.}
\label{fig:mesh_decimation}
\end{figure}

To train our method, we use a physics simulation as described in \secref{sec:simulation_stup}. We chose RaiSim \cite{hwangbo2018raisim}, since it allows modeling non-convex meshes and efficient parallel training. We first create a controllable hand model (\figref{fig:raisim}). Similar to \cite{yuan2021simpoe}, we compute the argmax of the skinning weights to assign each of the vertices to a body part. We then group the vertices accordingly and create a mesh for each body part. We limit the joint range in a data-driven manner. Specifically, we estimate the joint limits by parsing the DexYCB dataset and acquiring the maximum joint range, similar to \cite{spurr2020weakly}. Since the data may not contain the full range of possible joint displacements, we increase this limit by a slack constant. In practice, we found that approximating the collision bodies with primitive shapes (i.e., the simple objects and the hand meshes) led to an order of magnitude increase in training speed. This is because the simulation time increases roughly quadratically with the number of collision points. Therefore, for more complex object meshes, we apply a decimation technique to reduce the number of vertices (\figref{fig:mesh_decimation}). For the simpler meshes, we use primitive shapes and mesh alignment as an approximation. For training and evaluation, we therefore use the simplified meshes (except for the interpenetration metric, see \secref{app:metrics}).

\subsection{Learning Algorithm}

We train policies by using our own implementation of the widely used PPO algorithm \cite{schulman2017proximal}. We use the parameters summarized in \tabref{tab:ppo} for training. We create a parallelized training scheme with a worker per grasp label for data gathering  (amounting to e.g. 376 parallel environments for DexYCB). We then train a single policy over all objects, containing all grasps from the training set. For the GraspTTA \cite{jiang2021graspTTA} and ContactOpt \cite{grady2021contactopt} experiments, we double the amount of workers, such that they roughly correspond to the batch size of the DexYCB experiment (i.e., 400 workers with 2 workers for each label). Each training cycle utilizes a single GPU and 100 CPU cores and takes up to 24-72 hours of training. 

\begin{table}[ht]
\centering
\resizebox{0.7\columnwidth}{!}{
\begin{tabular}{ll}
\textbf{Hyperparameters PPO} & \textbf{Value} \vspace{0.1cm}\\
\hline\\
Epochs & 1e4\\
Steps per epoch &1.2e6\\
Environment steps for grasping & 195 \\
Environment steps for full task & 300 \\
Batch size & 376 \\
Updates per epoch & 16 \\
Simulation timestep & 2.22e-3s \\
Simulation steps per action & 13 \\
Discount factor $\gamma$ & 0.996 \\
GAE parameter $\lambda$ & 0.95 \\
Clipping parameter &0.2 \\
Max. gradient norm & 0.5 \\
Value loss coefficient & 0.5\\
Entropy coefficient & 0.0\\
Optimizer & Adam\cite{adam2015}\\
Learning rate & 5e-4\\
Hidden units & 128 \\
Hidden layers & 2 \\
\\
\textbf{Weight Parameters} & \textbf{Value} \vspace{0.1cm}\\
\hline\\
$w_{x}$ & -2.0\\
$w_{q}$ & -0.1\\
$w_{c}$ & 1.0\\
$w_{\text{reg},h}$ & 0.5\\
$w_{\text{reg},o}$ & 1.0 \\
$w_{x,j}$ & 1.0 \\
$w_{x,\text{tip}}$ & 4.0 \\
$\lambda$ & 5.0 \\
$\alpha_x$ & -2.0\\
$\alpha_q$ & -0.25\\

\end{tabular}
}
\captionof{table}{Hyperparameters of our method. The parameter "steps per epoch" is reported for the DexYCB training set with a batch size of 376. This number varies according on the amount of grasp labels available in the training set.}
\label{tab:ppo}
\end{table}

\subsection{Metrics Details}
\label{app:metrics}
This section contains an extended description of the metrics depicted in \secref{exp_metrics}. \\ 
\textbf{Success Rate:} We define the success rate as the primary measure of physical plausibility. 
It is measured as the rate of sequences which maintain a stable grasp, i.e., where the object does not slip and fall down for a period of a 5s window. We lower the surface in the simulation for this purpose. A success rate of 0.0 indicates no success, 1.0 means all sequences were successful.   \\
\textbf{Interpenetration:} We calculate the amount of hand volume that penetrates the object. To do so, we use the original MANO mesh \cite{romero2017mano} and the high-resolution object mesh. Hence, there is no physical simulation involved when measuring interpenetration. To ensure a fair comparison against the static baseline, we choose the last time step of the grasping phase for our method and hence omit the approaching phase from the evaluation. \\ 
\textbf{Simulated Distance:} Similar to the metric proposed in \cite{jiang2021graspTTA}, we compute the mean displacement between the object and the hand's wrist. Instead of measuring the absolute displacement, we report the mean displacement in mm per second. We measure the displacement for a maximum window of 5s or stop whenever the object falls and hits the surface.\\
\textbf{Contact Ratio:} For the ablation study, we measure the ratio between the target contacts $\overline{\mathbf{g}}_c$ defined via the grasp label $\grasplabel$ and the contacts achieved in the physics simulation $\mathbb{I}[\mathbf{f}>0]$. We average over the whole sequence, therefore both the approaching and grasping phase are contained in this metric. \\
\textbf{MPE:} This metric is used for the motion synthesis experiments. It is the mean position error between the object's 3D position and the object's target 3D position, defined as $\goalvec_{o,x}$ (\secref{app_feat_grasp}):
\begin{equation}
    e_{\text{mpe}} = \| \goalvec_{o,x} \|^2
    \label{eq_mpe}
\end{equation}
\textbf{Geodesic:} This is the angular metric used in the motion synthesis experiments. In particular, the angular distance between the object's \update{current} orientation $\posesix_{o,q}$ and the object's target orientation $\posesix_{g,q}$. It is defined as follows:

\begin{equation}
    e_{\text{geo}} = \text{acos}(0.5(\text{trace}(\mathbf{R}_o \mathbf{R}_g^\top)-1)),
        \label{eq_geo}
\end{equation}
where $\mathbf{R}_o$ and $\mathbf{R}_g$ are the rotation matrices of the corresponding orientations of the object and the target 6D pose, respectively.
\subsection{Baselines}

Here we provide an extended description of the baselines.\\
\textbf{*-PD:}
Similar to \cite{jiang2021graspTTA}, we place the object into the hand via the grasp label. We then attempt to maintain the grasp using PD-control in the physics simulation. To do so, the hand's 6DoF global pose $\textbf{T}_h$ and the joint configuration $\textbf{q}_h$ are initialized with the grasp label reference directly, hence $\textbf{T}_h=\overline{\textbf{T}}_h$ and $\textbf{q}_h=\overline{\textbf{q}}_h$. \\
\textbf{*-IK:}
We employ an offline optimization to correct for imperfections (i.e., minor distances or penetrations) in the label by utilizing the information about the target contacts $\overline{\mathbf{g}}_c$ (\secref{app_feat_grasp}) and the closest points on the object surface. In particular, for the finger parts that we deem to be in contact, we replace the original 3D keypoints from the grasp label $\overline{\mathbf{x}}_h$ by the closest vertex points on the object surface. We then run an optimization to yield a corrected target pose. The reconstructed samples are then passed to the PD-control. We found this technique to be effective for motion capture data, but not for the labels from GraspTTA \cite{jiang2021graspTTA} or ContactOpt \cite{grady2021contactopt}, likely because both methods already inherently optimize for contact. Hence, we omit it for the latter methods in the main text. \\
\textbf{Flat-RL:} 
 We employ an RL baseline that does not separate the grasping from the motion synthesis phase, but trains the full dynamic grasp synthesis task end-to-end. In particular, this baseline uses the concatenation of the grasping policy's feature layer $\phi(\statevec,\grasplabel)$ (\secref{method_feat_grasp}) and the feature layer of the learned motion synthesis module $\psi(\statevec,\posesix_g, \grasplabel)$ (\ref{app:motion_synthesis_feat}). Hence, the policy in this case is $\boldsymbol{\pi}(\actions|\phi(\statevec,\grasplabel),\psi(\statevec,\posesix_g, \grasplabel))$. For the reward function we use the combination of the reward used for the grasping policy (Eq. 4 in main paper) and the reward for the decoupled motion synthesis policy (Eq. \ref{rew_motion}). The weights of the different reward components are reported in \tabref{tab:ppo}. \\

\subsection{Experimental Details}

Here we provide a short overview of the different object sets and grasp labels used in each experiment.
\paragraph{Grasping Objects}
\label{sec:app_grasping_objects}
When using grasp predictions from an external grasp synthesis method \cite{jiang2021graspTTA} (Section \ref{exp_grasp}), we train with the objects used in DexYCB \cite{chao2021dexycb}. During evaluation, we report results on both the HO3D subset as done in \cite{jiang2021graspTTA} and the objects from DexYCB. For the experiment with ContactOpt \cite{grady2021contactopt}, we train and test on the HO3D objects (except for 019 pitcher base, which is not contained in the dataset). Note that since the models for grasp synthesis and the image-based pose estimates have no notion of physics in terms of where an object is positioned in space (in contrast to the data from DexYCB), we apply a small modification to the simulation to ensure a fair comparison. We place the object on a surface and allow the hand to approach from any direction, even penetrating the surface. We achieve this by disabling the collision response between the surface and the hand. In future work, an optimization could filter out poses that require approaching from beneath a surface. Also note that since we only have access to a single grasp reference and not a sequence for GraspTTA and ContactOpt, we start each sequence at a predefined distance away from the object in the mean MANO hand pose. \\
For the evaluation of our method in this experiment, we remove the surface (i.e. table) after the \emph{grasping} phase. The metrics are being measured from the moment the table is removed. For the baselines, we directly start the sequence in the target pose of both the hand and object (without a table present).

\paragraph{Motion Synthesis} For the experiment presented in \secref{exp_motion}, we included a representative subset of YCB\cite{calli2015ycb} objects. Namely, we used 2 cylindric objects (002 master chef can and 007 tuna can), 2 box-shaped objects (004 sugarbox and 061 foam), and 2 more complex objects (019 pitcher base and 052 extra large clamp) for training and evaluation. We use our train-split of DexYCB in this experiment. Furthermore, we filter out the failed grasps from the experiment in \secref{exp_grasp} and train and evaluate only on the stable grasps. Using unsuccessful grasps in this case would not produce any viable motions, since the objects cannot be grasped correctly to initiate the \emph{motion synthesis} phase. Each sequence starts with the \emph{grasping} phase, where only the grasping policy $\gpolicyvec$ is active. This ensures that a stable grasp on the object can be reached before moving the object globally. In the subsequent \emph{motion synthesis} phase, both the grasping policy and the motion synthesis module are acting simultaneously. 

\paragraph{Ablations} For the experiment presented in \secref{exp_ablation}, we use a subset of YCB\cite{calli2015ycb} objects and train per-object policies with grasp labels extracted from DexYCB\cite{chao2021dexycb}. In particular, we included one cylindric object (002 master chef can), one box-shaped object (004 sugarbox) and one complex object (052 extra large clamp) for training and evaluation.

\section{Additional Results}

\begin{table}[t]
	\centering
	\resizebox{\columnwidth}{!}{%
		\begin{tabular}{ll|ccc}
		
			\toprule
            
			& Models & Success $\uparrow$ & SimDist [mm/s] $\downarrow$ & Interpenetration $\downarrow$ \\ \midrule
			
			\parbox[t]{2mm}{\multirow{2}{*}{\rotatebox[origin=c]{90}{Train}}}
            &  Ours PO & \textbf{0.81}  & $\mathbf{3.7 \pm 5.8}$ & 1.94 \\
            & \greycellcol Ours AO & \greycellcol 0.7  & \greycellcol $5.8 \pm 7.4$ &  \greycellcol \textbf{1.75} \\ \midrule

			\parbox[t]{2mm}{\multirow{2}{*}{\rotatebox[origin=c]{90}{Test}}}
            & Ours PO & 0.42 & $12.4 \pm 10.4$ & \textbf{1.19} \\
            & \greycellcol Ours AO & \greycellcol \textbf{0.63}  & \greycellcol $\mathbf{8.0 \pm 8.1}$ &  \greycellcol 1.77 \\
			\bottomrule
		\end{tabular}%
	}

	\caption{\textbf{Policy Type Comparison}. We compare a single policy trained over multiple objects (Ours AO) and single policies trained per object (Ours PO). We find that the all-object policies lead to better generalization performance on the DexYCB dataset \cite{chao2021dexycb}.}
	\label{tab:app_policies}
\end{table}

\textbf{All-Object vs. Per-Object Policies} We experimented both with a single policy trained over multiple objects and single policies trained per object. A comparison on the DexYCB train/test-split is shown in Table \ref{tab:app_policies}. We observe that the single object policies (Ours PO) can be trained faster (i.e. $\sim$3000 vs. 10'000 epochs) and yield a better overall performance on the training labels, likely due to overfitting. On the other hand, the more general all-object policies (Ours AO) take longer to train, however, the generalization performance on unseen grasp labels is better. The performance on the training data is lower compared to the per-object policies. This result indicates that an all-object policy helps to prevent the policy from overfitting to single grasp references. \\

\begin{figure}
\begin{center}
  \includegraphics[width=0.9\columnwidth]{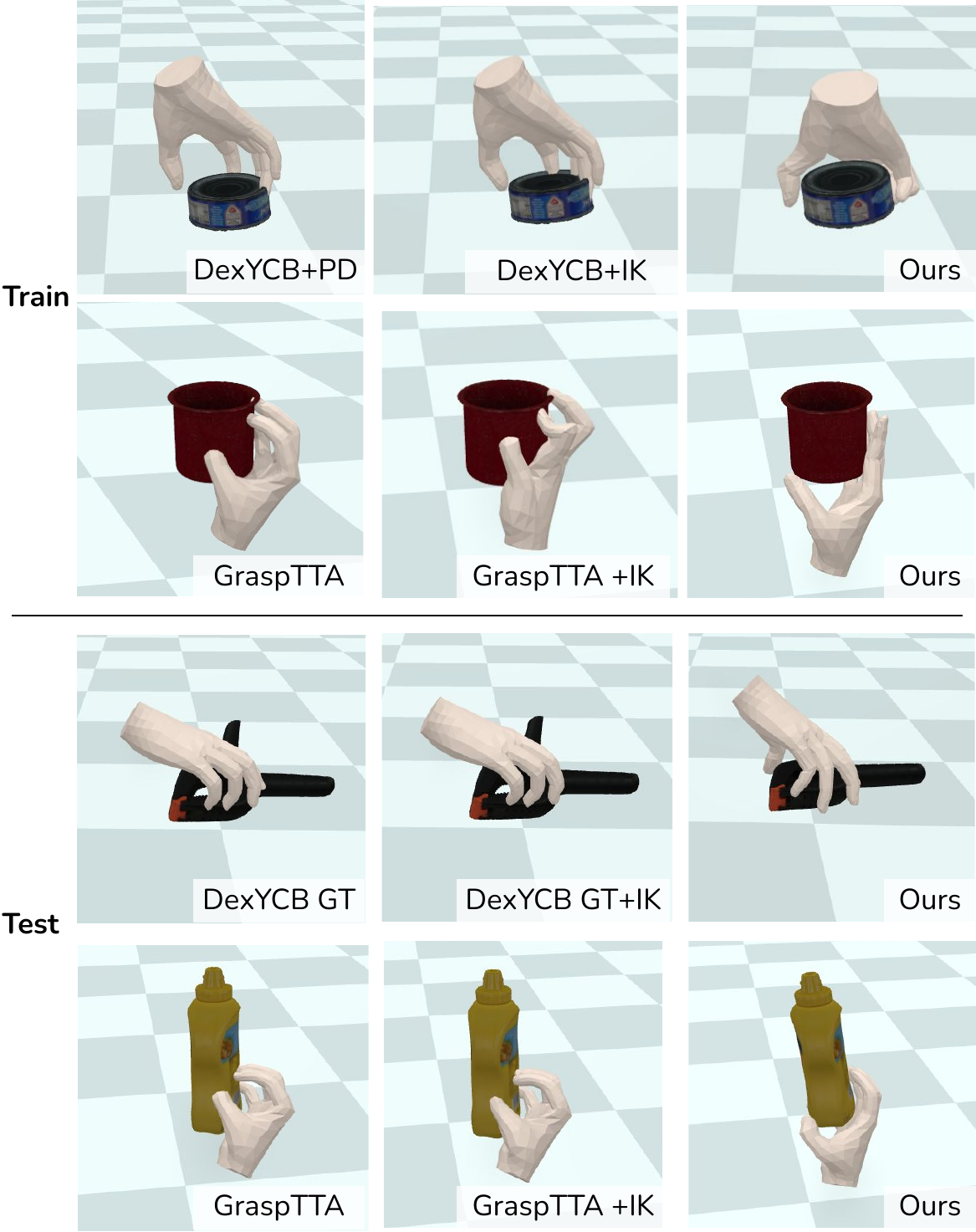}
\end{center}
\vspace{-0.4cm}
  \caption{\textbf{Additional Qualitative Grasps.} We provide additional qualitative examples of grasps. Rows 1-2: Comparison of the grasps on the training-sets of DexYCB \cite{chao2021dexycb} and the generated grasps from \cite{jiang2021graspTTA}. Rows 3-4: Comparison on the test-sets of DexYCB \cite{chao2021dexycb} and the generated grasps from \cite{jiang2021graspTTA}. As shown, our method produces more physically plausible grasps, i.e., with less interpenetration and more realistic contacts than the baselines.}
\label{fig:long}
\label{fig:app_qual_grasps}
\end{figure}

\textbf{Additional Qualitative Grasping Results}
We provide additional qualitative results in \figref{fig:app_qual_grasps}. Specifically, we include examples on the training sets of DexYCB \cite{chao2021dexycb} and the generated grasps \cite{jiang2021graspTTA}. Moreover, we present additional examples for both test-sets. As can be observed, our method can correct for interpenetration and achieve more realistic grasps. \\

\label{app_results}
\textbf{Quantitative Grasping Result Details}
We present the results of the empirical evaluation per object in Tables \ref{tab:exp1_train_full}-\ref{tab:exp1_test_contactopt_full}. It allows us to analyze the results in more detail.
For the grasp evaluation experiment (\secref{exp_grasp}), we find that the main difficulty for our learned policy are thin objects which are hard to pick up from the surface, e.g., grasping a pair of scissors from a table. This is indicated by the relatively low success rates in Tables \ref{tab:exp1_train_full} and \ref{tab:exp1_test_full} for the "037 scissors" and "040 large marker" objects. Grasping these objects requires very fine-grained  finger motion or creating a distinct motion to pick them up, which involves sliding the object along the surface to overcome static friction. We find that this issue is mitigated partially in the experiment with generated grasp labels (Tables \ref{tab:exp1_train_gtta_full} and \ref{tab:exp1_test_gtta_full}), because the deactivated collisions of the hand with the surface (see \secref{sec:app_grasping_objects}) help to achieve stable grasps. \\
For the baselines, we occasionally observe a configuration that leads to high success rates despite noisy pose references. Specifically, if the interpenetration is large (e.g. GT-IK in Table \ref{tab:exp1_test_full} for "021 bleach cleanser" or "024 bowl"), the objects can become entangled within the hand mesh and will therefore not be able to fall down. Thus, the success rate metric should always be interpreted in combination with the other metrics. \\
For the experiment with HO3D images, we find that the performance of our method is equally good across all objects and conditions (Tables \ref{tab:exp1_train_contactopt_full} and \ref{tab:exp1_test_contactopt_full}). This is likely due to the high-quality reference grasps that are produced by \cite{grady2021contactopt}.
While our approach can correct interpenetration and noisy poses to some degree, it is conditioned on the reference pose, and hence performs best when provided with grasp targets that roughly approximate a real human (i.e., physically plausible) grasp on the object. We conclude that especially for generalization to unseen objects, good grasp references are important. \\

\textbf{Generalization to Unseen Object Details}
In Table \ref{tab:app_generalization}, we report the detailed results of the generalization experiment. We observe a large variance across the different object sets. For example, the success rate of our method on test set 1, which comprises easier geometries, reaches up to 0.83. On the other hand, our method only achieves a 0.33 success rate on the test set 6, which contains the complex objects "037 scissors" and "040 large marker". Generally, we find that our method is able to outperform the static baselines across the different test sets. As a future extension, it would be interesting to scale the method to even larger datasets. Such ambitions are supported by different works for dexterous robotic manipulation tasks \cite{chen2021simple, huang2021geometry}, which have recently demonstrated the ability of large scale training with regards to object types in order to achieve generalization across objects. 


\section{Societal Impact}
\label{app_lim}
While the dynamic grasps generated by our method are not yet indistinguishable from real ones, we can extrapolate to a more mature version of this work, opening-up many potential applications, e.g., in AR/VR, HCI or robotics. These applications may lead to negative societal impact, where so-called deep-fakes are the obvious nefarious use of such methods. However, it is also possible that due to the computational complexity and resulting real-world cost of implementing even positive applications, there may be negative implications for already underprivileged populations. For example, a service robot that may learn to cooperate with humans may not be affordable for many that have need for such advanced care technologies. 

In going forward with the development of technologies related to this paper, one must carefully balance the potential positive uses and the undesired side-effects. Since we have no control over whether such technologies will be developed at all, by whom and for which purposes, we argue that openly discussing the technical details, properties and limitations is one way to ensure that a) such technologies are well understood and therefore counter measures to nefarious use would be easier to implement and b) that as many individuals as possible can have access to related technologies. To this end we will release all source code for research purposes.

\section{Glossary}
\label{app_glossary}
We include a glossary in Tables \ref{tab:glossary_p1} and \ref{tab:glossary_p2} to provide an overview of the many notations used in this paper.

\begin{table*}[ht]
\begin{center}
\resizebox{0.8\linewidth}{!}{
    \begin{tabular}{@{}l|c|c|c|c|c|c|c|c|c}
        \toprule
            \multicolumn{1}{c}{} & \multicolumn{3}{c|}{GT+PD} & \multicolumn{3}{c|}{GT+IK} & \multicolumn{3}{c}{Ours} \\ \midrule
        Object  & SimDist [mm/s] $\downarrow$ & Success $\uparrow$  & Interp. [$cm^3$]   & SimDist [mm/s] $\downarrow$ & Success $\uparrow$ & Interp. [$cm^3$] & SimDist [mm/s] $\downarrow$ & Success $\uparrow$ &  Interp. [$cm^3$]\\ \midrule
002 master chef can & 18.0 $\pm$ 9.8 & 0.19 & 5.68 & 17.2 $\pm$ 10.0 & 0.20 & 12.96 & 1.5 $\pm$ 5.7 & 0.90 & 1.54 \\
\rowcolor{Gray}
003 cracker box & 18.2 $\pm$ 10.1 & 0.16 & 3.62 & 9.2 $\pm$ 10.7 & 0.47 & 9.16 & 8.4 $\pm$ 12.3 & 0.68 & 2.48 \\
004 sugar box & 15.6 $\pm$ 11.2 & 0.32 & 5.52 & 8.7 $\pm$ 9.7 & 0.50 & 11.64 & 0.1 $\pm$ 0.0 & 1.00 & 3.28 \\
\rowcolor{Gray}
005 tomato soup can & 12.1 $\pm$ 10.8 & 0.28 & 4.34 & 12.2 $\pm$ 10.1 & 0.33 & 10.72 & 1.4 $\pm$ 5.5 & 0.90 & 2.51 \\
006 mustard bottle & 4.4 $\pm$ 7.8 & 0.64 & 9.51 & 1.1 $\pm$ 1.5 & 0.76 & 16.50 & 3.2 $\pm$ 8.5 & 0.88 & 2.20 \\
\rowcolor{Gray}
007 tuna fish can & 17.4 $\pm$ 9.4 & 0.14 & 2.52 & 16.6 $\pm$ 9.6 & 0.19 & 5.26 & 1.8 $\pm$ 4.6 & 0.71 & 1.13 \\
008 pudding box & 15.1 $\pm$ 9.7 & 0.21 & 3.91 & 13.2 $\pm$ 11.2 & 0.39 & 7.19 & 2.4 $\pm$ 5.9 & 0.78 & 0.98 \\
\rowcolor{Gray}
009 gelatin box & 18.9 $\pm$ 9.4 & 0.15 & 2.23 & 18.3 $\pm$ 9.3 & 0.20 & 4.61 & 3.7 $\pm$ 8.6 & 0.85 & 0.98 \\
010 potted meat can & 13.6 $\pm$ 10.3 & 0.30 & 4.14 & 11.4 $\pm$ 10.0 & 0.39 & 8.81 & 1.8 $\pm$ 6.0 & 0.89 & 0.64 \\
\rowcolor{Gray}
011 banana & 16.7 $\pm$ 8.6 & 0.09 & 3.27 & 15.9 $\pm$ 10.0 & 0.20 & 4.97 & 5.9 $\pm$ 9.6 & 0.47 & 0.73 \\
019 pitcher base & 11.0 $\pm$ 11.1 & 0.40 & 7.47 & 11.2 $\pm$ 11.2 & 0.37 & 17.01 & 6.9 $\pm$ 10.8 & 0.68 & 3.16 \\
\rowcolor{Gray}
021 bleach cleanser & 5.1 $\pm$ 8.5 & 0.61 & 8.25 & 3.8 $\pm$ 7.0 & 0.61 & 15.64 & 0.2 $\pm$ 0.4 & 0.94 & 3.08 \\
024 bowl & 10.2 $\pm$ 10.7 & 0.41 & 3.15 & 7.3 $\pm$ 9.3 & 0.62 & 9.83 & 7.5 $\pm$ 10.7 & 0.62 & 2.08 \\
\rowcolor{Gray}
025 mug & 12.3 $\pm$ 10.3 & 0.35 & 3.24 & 9.7 $\pm$ 10.6 & 0.53 & 6.56 & 10.2 $\pm$ 12.2 & 0.59 & 1.61 \\
035 power drill & 0.8 $\pm$ 1.8 & 0.83 & 10.68 & 5.3 $\pm$ 9.0 & 0.65 & 15.76 & 6.4 $\pm$ 10.7 & 0.59 & 1.64 \\
\rowcolor{Gray}
036 wood block & 13.4 $\pm$ 11.6 & 0.42 & 7.31 & 11.3 $\pm$ 11.5 & 0.50 & 13.36 & 0.1 $\pm$ 0.0 & 1.00 & 3.56 \\
037 scissors & 10.6 $\pm$ 9.9 & 0.35 & 1.72 & 13.5 $\pm$ 9.4 & 0.22 & 3.03 & 19.3 $\pm$ 10.0 & 0.11 & 0.35 \\
\rowcolor{Gray}
040 large marker & 19.8 $\pm$ 5.8 & 0.04 & 1.38 & 21.5 $\pm$ 5.9 & 0.05 & 2.81 & 20.9 $\pm$ 7.3 & 0.05 & 0.09 \\
052 extra large clamp & 15.0 $\pm$ 9.9 & 0.28 & 1.97 & 12.3 $\pm$ 11.0 & 0.44 & 3.50 & 10.4 $\pm$ 11.7 & 0.44 & 1.70 \\
\rowcolor{Gray}
061 foam brick & 18.9 $\pm$ 7.3 & 0.12 & 1.93 & 17.2 $\pm$ 10.5 & 0.26 & 5.23 & 3.5 $\pm$ 8.1 & 0.84 & 1.30 \\
\midrule
Average & 13.4 $\pm$ 9.2 & 0.31 & 4.59 & 11.8 $\pm$ 9.4 & 0.39 & 9.23 & 5.8 $\pm$ 7.4 & 0.70 & 1.75 \\ \bottomrule
    \end{tabular}}
\end{center}
\vspace{-0.2cm}
\caption{Detailed results for the DexYCB train set.}
\label{tab:exp1_train_full}
\end{table*}

\begin{table*}[t]
\begin{center}
\resizebox{0.8\linewidth}{!}{
    \begin{tabular}{@{}l|c|c|c|c|c|c|c|c|c}
        \toprule
            \multicolumn{1}{c}{} & \multicolumn{3}{c|}{GT+PD} & \multicolumn{3}{c|}{GT+IK} & \multicolumn{3}{c}{Ours} \\ \midrule
        Object  & SimDist [mm/s] $\downarrow$ & Success $\uparrow$  & Interp. [$cm^3$]   & SimDist [mm/s] $\downarrow$ & Success  $\uparrow$ & Interp. [$cm^3$] & SimDist [mm/s] $\downarrow$ & Success $\uparrow$ &  Interp. [$cm^3$]\\ \midrule
002 master chef can & 20.4 $\pm$ 9.3 & 0.17 & 4.17 & 19.2 $\pm$ 8.6 & 0.17 & 10.73 & 0.6 $\pm$ 1.2 & 0.83 & 1.67 \\
\rowcolor{Gray}
003 cracker box & 14.7 $\pm$ 11.1 & 0.33 & 5.21 & 9.0 $\pm$ 11.5 & 0.67 & 13.10 & 8.9 $\pm$ 12.4 & 0.67 & 4.06 \\
004 sugar box & 14.1 $\pm$ 12.1 & 0.43 & 6.75 & 3.9 $\pm$ 8.5 & 0.86 & 16.04 & 3.9 $\pm$ 9.3 & 0.86 & 3.02 \\
\rowcolor{Gray}
005 tomato soup can & 7.6 $\pm$ 9.0 & 0.50 & 3.25 & 8.2 $\pm$ 8.4 & 0.25 & 7.56 & 14.4 $\pm$ 11.4 & 0.25 & 1.03 \\
006 mustard bottle & 9.5 $\pm$ 9.9 & 0.50 & 6.77 & 6.7 $\pm$ 9.9 & 0.63 & 13.19 & 7.0 $\pm$ 12.0 & 0.75 & 2.13 \\
\rowcolor{Gray}
007 tuna fish can & 16.4 $\pm$ 9.4 & 0.14 & 1.55 & 16.3 $\pm$ 8.5 & 0.14 & 3.55 & 0.1 $\pm$ 0.0 & 1.00 & 1.50 \\
008 pudding box & 18.5 $\pm$ 8.2 & 0.17 & 2.08 & 12.7 $\pm$ 10.5 & 0.33 & 4.19 & 4.2 $\pm$ 9.2 & 0.83 & 0.92 \\
\rowcolor{Gray}
009 gelatin box & 21.6 $\pm$ 9.7 & 0.14 & 2.00 & 15.3 $\pm$ 13.3 & 0.29 & 4.59 & 3.8 $\pm$ 8.5 & 0.71 & 0.86 \\
010 potted meat can & 9.9 $\pm$ 10.2 & 0.40 & 2.60 & 10.8 $\pm$ 10.8 & 0.40 & 7.00 & 0.1 $\pm$ 0.1 & 1.00 & 1.00 \\
\rowcolor{Gray}
011 banana & 13.8 $\pm$ 9.9 & 0.14 & 2.79 & 13.0 $\pm$ 10.8 & 0.29 & 4.36 & 10.8 $\pm$ 12.0 & 0.43 & 0.50 \\
019 pitcher base & 12.6 $\pm$ 12.5 & 0.50 & 7.83 & 13.1 $\pm$ 12.7 & 0.50 & 25.54 & 6.2 $\pm$ 9.1 & 0.50 & 3.08 \\
\rowcolor{Gray}
021 bleach cleanser & 5.8 $\pm$ 8.3 & 0.40 & 10.43 & 5.1 $\pm$ 9.3 & 0.80 & 15.40 & 10.2 $\pm$ 12.3 & 0.60 & 2.10 \\
024 bowl & 9.3 $\pm$ 11.1 & 0.50 & 5.83 & 0.3 $\pm$ 0.3 & 1.00 & 13.44 & 12.0 $\pm$ 11.9 & 0.50 & 1.56 \\
\rowcolor{Gray}
025 mug & 7.4 $\pm$ 9.2 & 0.50 & 3.00 & 0.5 $\pm$ 0.5 & 1.00 & 9.13 & 4.1 $\pm$ 9.0 & 0.83 & 1.33 \\
035 power drill & 0.2 $\pm$ 0.1 & 1.00 & 8.25 & 0.5 $\pm$ 0.4 & 0.83 & 15.02 & 5.6 $\pm$ 10.4 & 0.67 & 2.71 \\
\rowcolor{Gray}
036 wood block & 20.6 $\pm$ 9.1 & 0.17 & 4.88 & 12.9 $\pm$ 12.4 & 0.50 & 11.73 & 0.2 $\pm$ 0.1 & 1.00 & 4.17 \\
037 scissors & 8.1 $\pm$ 9.9 & 0.50 & 4.44 & 6.4 $\pm$ 8.6 & 0.50 & 6.55 & 19.0 $\pm$ 8.1 & 0.13 & 0.58 \\
\rowcolor{Gray}
040 large marker & 18.7 $\pm$ 3.2 & 0.00 & 1.43 & 13.5 $\pm$ 8.7 & 0.20 & 3.58 & 24.2 $\pm$ 0.1 & 0.00 & 0.00 \\
052 extra large clamp & 16.2 $\pm$ 8.9 & 0.14 & 2.48 & 5.1 $\pm$ 7.7 & 0.43 & 5.21 & 14.0 $\pm$ 12.1 & 0.43 & 1.88 \\
\rowcolor{Gray}
061 foam brick & 16.0 $\pm$ 10.8 & 0.29 & 2.43 & 10.0 $\pm$ 8.3 & 0.13 & 4.86 & 10.5 $\pm$ 12.0 & 0.57 & 1.30 \\  \midrule
Average & 13.1 $\pm$ 9.1 & 0.35 & 4.41 & 9.1 $\pm$ 8.5 & 0.50 & 9.74 & 8.0 $\pm$ 8.1 & 0.63 & 1.77 \\ \bottomrule
    \end{tabular}}
\end{center}
\caption{Detailed results for the DexYCB test set.}
\label{tab:exp1_test_full}
\end{table*}

\begin{table*}[t]
\begin{center}
\resizebox{0.8\linewidth}{!}{
    \begin{tabular}{@{}l|c|c|c|c|c|c|c|c|c}
        \toprule
            \multicolumn{1}{c}{} & \multicolumn{3}{c|}{Jiang \textit{et. al} \cite{jiang2021graspTTA}+PD} & \multicolumn{3}{c|}{Jiang \textit{et. al} \cite{jiang2021graspTTA}+IK} & \multicolumn{3}{c}{Ours} \\ \midrule
        Object  & SimDist [mm/s] $\downarrow$ & Success $\uparrow$  & Interp. [$cm^3$]   & SimDist [mm/s] $\downarrow$ & Success $\uparrow$ & Interp. [$cm^3$] & SimDist [mm/s] $\downarrow$ & Success $\uparrow$ &  Interp. [$cm^3$]\\ \midrule
002 master chef can & 24.0 $\pm$ 1.5 & 0.00 & 6.39 & 24.0 $\pm$ 1.6 & 0.00 & 16.19 & 2.6 $\pm$ 7.5 & 0.90 & 2.70 \\
\rowcolor{Gray}
003 cracker box* & 22.8 $\pm$ 2.4 & 0.00 & 6.97 & 22.8 $\pm$ 2.5 & 0.00 & 8.18 & 6.0 $\pm$ 10.3 & 0.70 & 5.80 \\
004 sugar box* & 14.7 $\pm$ 8.5 & 0.10 & 5.60 & 15.2 $\pm$ 8.7 & 0.10 & 10.45 & 1.4 $\pm$ 5.3 & 0.95 & 4.03 \\
\rowcolor{Gray}
005 tomato soup can & 14.8 $\pm$ 9.1 & 0.10 & 5.62 & 14.6 $\pm$ 9.0 & 0.10 & 11.51 & 3.8 $\pm$ 8.7 & 0.85 & 5.76 \\
006 mustard bottle* & 5.6 $\pm$ 7.2 & 0.30 & 5.66 & 5.2 $\pm$ 7.3 & 0.50 & 10.61 & 0.8 $\pm$ 2.8 & 0.95 & 5.14 \\
\rowcolor{Gray}
007 tuna fish can & 20.9 $\pm$ 1.7 & 0.00 & 2.88 & 21.4 $\pm$ 1.7 & 0.00 & 4.84 & 0.3 $\pm$ 0.5 & 0.90 & 2.03 \\
008 pudding box & 11.0 $\pm$ 10.3 & 0.10 & 4.92 & 10.9 $\pm$ 10.2 & 0.10 & 7.88 & 6.2 $\pm$ 10.1 & 0.70 & 0.91 \\
\rowcolor{Gray}
009 gelatin box & 14.1 $\pm$ 8.7 & 0.10 & 3.89 & 14.3 $\pm$ 9.2 & 0.10 & 8.38 & 1.7 $\pm$ 5.9 & 0.90 & 1.44 \\
010 potted meat can* & 20.4 $\pm$ 2.9 & 0.00 & 3.96 & 20.7 $\pm$ 2.8 & 0.00 & 6.89 & 7.1 $\pm$ 11.0 & 0.65 & 0.46 \\
\rowcolor{Gray}
011 banana* & 4.4 $\pm$ 6.1 & 0.30 & 3.52 & 6.1 $\pm$ 8.0 & 0.40 & 2.33 & 7.0 $\pm$ 9.4 & 0.45 & 3.18 \\
019 pitcher base* & 6.4 $\pm$ 8.4 & 0.50 & 8.11 & 7.1 $\pm$ 8.0 & 0.30 & 13.64 & 6.4 $\pm$ 10.1 & 0.65 & 1.56 \\
\rowcolor{Gray}
021 bleach cleanser* & 0.8 $\pm$ 1.9 & 0.90 & 5.76 & 0.5 $\pm$ 0.6 & 0.80 & 6.82 & 2.1 $\pm$ 6.4 & 0.90 & 4.89 \\
024 bowl & 5.5 $\pm$ 8.0 & 0.70 & 4.93 & 5.7 $\pm$ 8.3 & 0.70 & 3.24 & 5.2 $\pm$ 7.7 & 0.40 & 1.78 \\
\rowcolor{Gray}
025 mug* & 7.9 $\pm$ 9.4 & 0.50 & 4.32 & 8.2 $\pm$ 9.5 & 0.40 & 2.19 & 1.3 $\pm$ 5.2 & 0.95 & 4.78 \\
035 power drill* & 12.3 $\pm$ 12.0 & 0.20 & 5.84 & 12.1 $\pm$ 12.1 & 0.20 & 8.40 & 11.4 $\pm$ 11.5 & 0.20 & 1.27 \\
\rowcolor{Gray}
036 wood block & 22.0 $\pm$ 4.6 & 0.00 & 7.28 & 22.2 $\pm$ 4.2 & 0.00 & 5.06 & 2.6 $\pm$ 7.5 & 0.90 & 2.99 \\
037 scissors* & 5.0 $\pm$ 7.5 & 0.30 & 2.37 & 8.2 $\pm$ 8.5 & 0.30 & 1.50 & 0.7 $\pm$ 1.4 & 0.85 & 2.18 \\
\rowcolor{Gray}
040 large marker & 10.3 $\pm$ 9.6 & 0.40 & 1.86 & 10.1 $\pm$ 9.7 & 0.40 & 3.41 & 7.2 $\pm$ 9.6 & 0.55 & 0.92 \\
052 extra large clamp & 3.6 $\pm$ 5.9 & 0.40 & 4.97 & 3.6 $\pm$ 6.0 & 0.40 & 8.21 & 3.4 $\pm$ 7.2 & 0.65 & 3.37 \\
\rowcolor{Gray}
061 foam brick & 20.9 $\pm$ 1.8 & 0.00 & 3.46 & 21.3 $\pm$ 2.0 & 0.00 & 6.56 & 1.3 $\pm$ 5.1 & 0.95 & 1.66 \\ \midrule
Average & 12.4 $\pm$ 6.4 & 0.25 & 4.92 & 12.7 $\pm$ 6.5 & 0.24 & 7.31 & 3.9 $\pm$ 7.2 & 0.75 & 2.84 \\ \bottomrule
    \end{tabular}}
\end{center}
\caption{Detailed results for the DexYCB and HO3D train set with grasp references from a static grasp synthesis method \cite{jiang2021graspTTA}. HO3D objects are marked by *.}
\label{tab:exp1_train_gtta_full}
\end{table*}
\begin{table*}[t]
\begin{center}
\resizebox{0.8\linewidth}{!}{
    \begin{tabular}{@{}l|c|c|c|c|c|c|c|c|c}
        \toprule
            \multicolumn{1}{c}{} & \multicolumn{3}{c|}{Jiang \textit{et. al} \cite{jiang2021graspTTA}+PD} & \multicolumn{3}{c|}{Jiang \textit{et. al} \cite{jiang2021graspTTA}+IK} & \multicolumn{3}{c}{Ours} \\ \midrule
        Object  & SimDist [mm/s] $\downarrow$ & Success $\uparrow$  & Interp. [$cm^3$]   & SimDist [mm/s] $\downarrow$ & Success $\uparrow$ & Interp. [$cm^3$] & SimDist [mm/s] $\downarrow$ & Success $\uparrow$ &  Interp. [$cm^3$]\\ \midrule
002 master chef can & 24.0 $\pm$ 1.6 & 0.00 & 7.54 & 23.3 $\pm$ 1.3 & 0.00 & 7.03 & 5.3 $\pm$ 10.5 & 0.80 & 1.88 \\
\rowcolor{Gray}
003 cracker box* & 22.8 $\pm$ 2.5 & 0.00 & 6.73 & 20.3 $\pm$ 5.4 & 0.00 & 7.74 & 5.4 $\pm$ 10.6 & 0.80 & 7.06 \\
004 sugar box* & 15.2 $\pm$ 8.7 & 0.10 & 5.18 & 15.0 $\pm$ 8.4 & 0.00 & 13.85 & 0.2 $\pm$ 0.2 & 1.00 & 3.48 \\
\rowcolor{Gray}
005 tomato soup can & 14.6 $\pm$ 9.0 & 0.10 & 5.29 & 15.0 $\pm$ 9.0 & 0.00 & 10.75 & 0.1 $\pm$ 0.1 & 1.00 & 5.86 \\
006 mustard bottle* & 5.2 $\pm$ 7.3 & 0.50 & 5.19 & 7.6 $\pm$ 7.6 & 0.10 & 14.29 & 0.4 $\pm$ 0.9 & 0.90 & 6.23 \\
\rowcolor{Gray}
007 tuna fish can & 21.4 $\pm$ 1.7 & 0.00 & 2.61 & 19.8 $\pm$ 6.7 & 0.10 & 5.25 & 2.5 $\pm$ 7.0 & 0.90 & 1.51 \\
008 pudding box & 10.9 $\pm$ 10.2 & 0.10 & 5.51 & 7.0 $\pm$ 9.1 & 0.30 & 9.38 & 7.3 $\pm$ 10.7 & 0.60 & 0.46 \\
\rowcolor{Gray}
009 gelatin box & 14.3 $\pm$ 9.2 & 0.10 & 3.49 & 13.0 $\pm$ 10.6 & 0.40 & 6.91 & 0.8 $\pm$ 1.6 & 0.80 & 1.83 \\
010 potted meat can* & 20.7 $\pm$ 2.8 & 0.00 & 4.74 & 21.0 $\pm$ 2.6 & 0.00 & 8.70 & 7.6 $\pm$ 10.4 & 0.40 & 0.60 \\
\rowcolor{Gray}
011 banana* & 6.1 $\pm$ 8.0 & 0.40 & 3.38 & 8.1 $\pm$ 8.9 & 0.50 & 3.16 & 4.8 $\pm$ 9.1 & 0.80 & 2.09 \\
019 pitcher base* & 7.1 $\pm$ 8.0 & 0.30 & 8.50 & 6.3 $\pm$ 8.1 & 0.40 & 0.00 & 12.3 $\pm$ 11.9 & 0.40 & 0.98 \\
\rowcolor{Gray}
021 bleach cleanser* & 0.5 $\pm$ 0.6 & 0.80 & 6.50 & 9.9 $\pm$ 9.1 & 0.30 & 7.75 & 0.2 $\pm$ 0.2 & 1.00 & 5.69 \\
024 bowl & 5.7 $\pm$ 8.3 & 0.70 & 4.51 & 2.5 $\pm$ 5.1 & 0.80 & 2.83 & 3.4 $\pm$ 7.0 & 0.80 & 2.18 \\
\rowcolor{Gray}
025 mug* & 8.2 $\pm$ 9.5 & 0.40 & 5.94 & 3.7 $\pm$ 6.3 & 0.60 & 2.41 & 0.1 $\pm$ 0.0 & 1.00 & 4.64 \\
035 power drill* & 12.1 $\pm$ 12.1 & 0.20 & 4.91 & 14.7 $\pm$ 12.0 & 0.30 & 5.91 & 10.2 $\pm$ 11.4 & 0.20 & 1.49 \\
\rowcolor{Gray}
036 wood block & 22.2 $\pm$ 4.2 & 0.00 & 6.65 & 24.0 $\pm$ 2.0 & 0.00 & 1.94 & 10.7 $\pm$ 12.4 & 0.50 & 1.84 \\
037 scissors* & 8.2 $\pm$ 8.5 & 0.30 & 2.94 & 5.1 $\pm$ 7.1 & 0.50 & 1.26 & 7.8 $\pm$ 11.3 & 0.60 & 1.74 \\
\rowcolor{Gray}
040 large marker & 10.1 $\pm$ 9.7 & 0.40 & 1.65 & 6.8 $\pm$ 9.9 & 0.60 & 4.01 & 7.3 $\pm$ 10.4 & 0.50 & 1.94 \\
052 extra large clamp & 3.6 $\pm$ 6.0 & 0.40 & 4.09 & 3.1 $\pm$ 6.1 & 0.60 & 7.46 & 5.1 $\pm$ 8.4 & 0.60 & 3.22 \\
\rowcolor{Gray}
061 foam brick & 21.3 $\pm$ 2.0 & 0.00 & 3.51 & 20.9 $\pm$ 1.3 & 0.00 & 7.81 & 0.1 $\pm$ 0.0 & 1.00 & 1.46 \\ \midrule
Average & 12.7 $\pm$ 6.5 & 0.24 & 4.94 & 12.4 $\pm$ 6.8 & 0.28 & 6.42 & 4.6 $\pm$ 6.7 & 0.73 & 2.81 \\ \bottomrule
    \end{tabular}}
\end{center}
\caption{Detailed results for the DexYCB and HO3D test set with grasp references from a static grasp synthesis method \cite{jiang2021graspTTA}. HO3D objects are marked by *.}
\label{tab:exp1_test_gtta_full}
\end{table*}
\begin{table*}[t]
\begin{center}
\resizebox{0.8\linewidth}{!}{
    \begin{tabular}{@{}l|c|c|c|c|c|c}
        \toprule
            \multicolumn{1}{c}{} & \multicolumn{3}{c|}{Grady \textit{et. al} \cite{grady2021contactopt}+PD} & \multicolumn{3}{c}{Ours} \\ \midrule
        Object  & SimDist [mm/s] $\downarrow$ & Success $\uparrow$  & Interp. [$cm^3$]   & SimDist [mm/s] $\downarrow$ & Success $\uparrow$ &  Interp. [$cm^3$]\\ \midrule
\rowcolor{Gray}
003 cracker box & 2.5 $\pm$ 6.7 & 0.85 & 14.33 & 0.3 $\pm$ 0.1 & 1.00 & 3.18 \\
004 sugar box & 16.3 $\pm$ 9.3 & 0.05 & 17.04 & 2.9 $\pm$ 6.6 & 0.70 & 2.40 \\
\rowcolor{Gray}
006 mustard bottle & 9.1 $\pm$ 9.7 & 0.40 & 26.46 & 0.3 $\pm$ 0.4 & 0.95 & 2.89 \\
010 potted meat can & 3.9 $\pm$ 8.5 & 0.70 & 15.42 & 2.2 $\pm$ 5.9 & 0.90 & 0.78 \\
\rowcolor{Gray}
011 banana & 10.0 $\pm$ 9.9 & 0.35 & 13.80 & 1.7 $\pm$ 4.9 & 0.80 & 1.98 \\
021 bleach cleanser & 0.9 $\pm$ 3.3 & 0.95 & 18.84 & 0.3 $\pm$ 0.1 & 1.00 & 2.86 \\
\rowcolor{Gray}
025 mug & 2.7 $\pm$ 6.7 & 0.80 & 5.49 & 2.0 $\pm$ 5.4 & 0.85 & 4.74 \\
035 power drill & 0.2 $\pm$ 0.4 & 0.95 & 16.09 & 0.3 $\pm$ 0.2 & 1.00 & 2.56 \\
\rowcolor{Gray}
037 scissors & 0.1 $\pm$ 0.1 & 1.00 & 6.96 & 3.0 $\pm$ 7.3 & 0.75 & 2.60 \\
\midrule
Average & 5.1 $\pm$ 6.1 & 0.67 & 14.94 & 1.4 $\pm$ 3.4 & 0.88 & 2.67 \\ \bottomrule
    \end{tabular}}
\end{center}
\caption{Detailed results for the train set with ContactOpt \cite{grady2021contactopt} on HO3D images.}
\label{tab:exp1_train_contactopt_full}
\end{table*}
\begin{table*}[t]
\begin{center}
\resizebox{0.8\linewidth}{!}{
    \begin{tabular}{@{}l|c|c|c|c|c|c}
        \toprule
            \multicolumn{1}{c}{} & \multicolumn{3}{c|}{Grady \textit{et. al} \cite{grady2021contactopt}+PD} & \multicolumn{3}{c}{Ours} \\ \midrule
        Object  & SimDist [mm/s] $\downarrow$ & Success $\uparrow$  & Interp. [$cm^3$]   & SimDist [mm/s] $\downarrow$ & Success $\uparrow$ &  Interp. [$cm^3$]\\ \midrule
\rowcolor{Gray}
003 cracker box & 6.7 $\pm$ 9.7 & 0.60 & 13.41 & 0.26 $\pm$ 0.1 & 1.00 & 2.14 \\
004 sugar box & 23.6 $\pm$ 1.5 & 0.00 & 17.71 & 0.95 $\pm$ 2.2 & 0.90 & 2.53 \\
\rowcolor{Gray}
006 mustard bottle & 4.6 $\pm$ 7.6 & 0.60 & 25.16 & 0.64 $\pm$ 0.8 & 0.80 & 2.46 \\
010 potted meat can & 1.9 $\pm$ 5.5 & 0.90 & 14.08 & 0.63 $\pm$ 1.3 & 0.90 & 0.38 \\
\rowcolor{Gray}
011 banana & 12.0 $\pm$ 10.4 & 0.20 & 13.38 & 0.61 $\pm$ 0.5 & 0.80 & 1.91 \\
021 bleach cleanser & 0.9 $\pm$ 2.4 & 0.90 & 18.23 & 2.86 $\pm$ 7.6 & 0.90 & 2.25 \\
\rowcolor{Gray}
025 mug & 6.4 $\pm$ 8.4 & 0.50 & 4.80 & 5.05 $\pm$ 9.7 & 0.80 & 4.16 \\
035 power drill & 0.1 $\pm$ 0.1 & 1.00 & 14.84 & 0.71 $\pm$ 0.6 & 0.60 & 1.68 \\
\rowcolor{Gray}
037 scissors & 2.7 $\pm$ 6.9 & 0.70 & 4.34 & 5.30 $\pm$ 9.3 & 0.60 & 1.21 \\
\midrule
Average & 6.5 $\pm$ 5.8 & 0.60 & 13.99 & 1.9 $\pm$ 3.57 & 0.81 & 2.08 \\ \bottomrule
    \end{tabular}}
\end{center}
\caption{Detailed results for the test set with ContactOpt \cite{grady2021contactopt} on HO3D images.}
\label{tab:exp1_test_contactopt_full}
\end{table*}
\newcommand{\gclr}{\cellcolor[HTML]{E4E4E4}}
\begin{table*}[ht]
\begin{center}
\resizebox{1.0\linewidth}{!}{
    \begin{tabular}{l|l|c|c|c|c|c|c|c|c|c}
        \toprule
         &   \multicolumn{1}{c}{} & \multicolumn{3}{c|}{GT+PD} & \multicolumn{3}{c|}{GT+IK} & \multicolumn{3}{c}{Ours} \\ \midrule
       & Object  & SimDist [mm/s] $\downarrow$ & Success $\uparrow$  & Interp. [$cm^3$]   & SimDist [mm/s] $\downarrow$ & Success $\uparrow$ & Interp. [$cm^3$] & SimDist [mm/s] $\downarrow$ & Success $\uparrow$ &  Interp. [$cm^3$]\\ \midrule

\parbox[t]{2mm}{\multirow{3}{*}{\rotatebox[origin=c]{90}{\scriptsize{Test set 1}}}} 
& \gclr 004 sugar box & \gclr 15.2 $\pm$ 11.4 & \gclr 0.35 & \gclr 5.86 & \gclr 7.4 $\pm$ 9.4 & \gclr 0.60 & \gclr 12.87 & \gclr 2.3 $\pm$ 7.2 & \gclr 0.92 & \gclr 3.35 \\ 
& 005 tomato soup can & 11.4 $\pm$ 10.5 & 0.32 & 4.17 & 11.5 $\pm$ 9.8 & 0.32 & 10.21 & 5.7 $\pm$ 10.3 & 0.76 & 1.56 \\
& \gclr 006 mustard bottle & \gclr 6.0 $\pm$ 8.5 & \gclr 0.60 & \gclr 8.64 & \gclr 2.9 $\pm$ 4.2 & \gclr 0.72 & \gclr 15.44 & \gclr 2.1 $\pm$ 5.7 & \gclr 0.80 & \gclr 1.87 \\ \midrule
& Average & 10.9 $\pm$ 10.1 & 0.42 & 6.22 & 7.3 $\pm$ 7.8 & 0.55 & 12.84 & 3.3 $\pm$ 7.7 & 0.83 & 2.26 \\ \midrule
& & & & & & & & & & \\

\parbox[t]{2mm}{\multirow{3}{*}{\rotatebox[origin=c]{90}{\scriptsize{Test set 2}}}} 
& \gclr 061 foam brick & \gclr 18.1 $\pm$ 8.3 & \gclr 0.16 & \gclr 2.06 & \gclr 15.3 $\pm$ 9.9 & \gclr 0.23 & \gclr 5.13 & \gclr 9.6 $\pm$ 12.0 & \gclr 0.62 & \gclr 0.55 \\
& 010 potted meat can & 12.8 $\pm$ 10.2 & 0.33 & 3.80 & 11.3 $\pm$ 10.2 & 0.39 & 8.42 & 5.1 $\pm$ 9.3 & 0.61 & 1.36 \\
& \gclr 052 extra large clamp & \gclr 15.4 $\pm$ 9.6 & \gclr 0.24 & \gclr 2.11 & \gclr 10.3 $\pm$ 10.1 & \gclr 0.44 & \gclr 3.98 & \gclr 14.8 $\pm$ 11.5 & \gclr 0.16 & \gclr 1.53 \\ \midrule
& Average & 15.4 $\pm$ 9.4 & 0.24 & 2.66 & 12.3 $\pm$ 10.1 & 0.35 & 5.84 & 9.8 $\pm$ 10.9 & 0.46 & 1.15 \\ \midrule
& & & & & & & & & & \\

\parbox[t]{2mm}{\multirow{3}{*}{\rotatebox[origin=c]{90}{\scriptsize{Test set 3}}}} 
& \gclr 003 cracker box & \gclr 17.4 $\pm$ 10.3 & \gclr 0.20 & \gclr 4.00 & \gclr 9.1 $\pm$ \gclr 10.9 & \gclr 0.52 & \gclr 10.11 & \gclr 12.9 $\pm$ 13.3 & \gclr 0.52 & \gclr 2.86 \\
& 007 tuna fish can & 17.1 $\pm$ 9.4 & 0.14 & 2.28 & 16.5 $\pm$ 9.3 & 0.18 & 4.83 & 5.4 $\pm$ 10.2 & 0.79 & 0.88 \\
& \gclr 011 banana & \gclr 15.8 $\pm$ 9.0 & \gclr 0.11 & \gclr 3.11 & \gclr 14.9 $\pm$ 10.2 & \gclr 0.23 & \gclr 4.78 & \gclr 7.2 $\pm$ 10.3 & \gclr 0.50 & \gclr 1.24 \\ \midrule
& Average & 16.8 $\pm$ 9.6 & 0.15 & 3.13 & 13.5 $\pm$ 10.2 & 0.31 & 6.57 & 8.5 $\pm$ 11.3 & 0.60 & 1.66 \\ \midrule
& & & & & & & & & & \\

\parbox[t]{2mm}{\multirow{3}{*}{\rotatebox[origin=c]{90}{\scriptsize{Test set 4}}}} 
& \gclr 002 master chef can & \gclr 18.5 $\pm$ 9.7 & \gclr 0.19 & \gclr 5.33 & \gclr 17.6 $\pm$ 9.7 & \gclr 0.19 & \gclr 12.45 & \gclr 6.6 $\pm$ 11.1 & \gclr 0.65 & \gclr 1.21 \\
& 036 wood block & 15.1 $\pm$ 11.1 & 0.36 & 6.75 & 11.7 $\pm$ 11.7 & 0.50 & 12.98 & 3.4 $\pm$ 8.8 & 0.85 & 3.31 \\
& \gclr 052 extra large clamp & \gclr 15.4 $\pm$ 9.6 & \gclr 0.24 & \gclr 2.11 & \gclr 10.3 $\pm$ 10.1 & \gclr 0.44 & \gclr 3.98 & \gclr 15.1 $\pm$ 11.3 & \gclr 0.28 & \gclr 2.26 \\ \midrule
& Average & 16.3 $\pm$ 10.1 & 0.26 & 4.73 & 13.2 $\pm$ 10.5 & 0.38 & 9.80 & 8.4 $\pm$ 10.4 & 0.59 & 2.26 \\ \midrule
& & & & & & & & & & \\

\parbox[t]{2mm}{\multirow{3}{*}{\rotatebox[origin=c]{90}{\scriptsize{Test set 5}}}} 
& \gclr 008 pudding box & \gclr 16.0 $\pm$ 9.4 & \gclr 0.20 & \gclr 3.45 & \gclr 13.0 $\pm$ 11.0 & \gclr 0.38 & \gclr 6.44 & \gclr 3.5 $\pm$ 8.4 & \gclr 0.83 & \gclr 0.90 \\
& 019 pitcher base & 11.4 $\pm$ 11.4 & 0.42 & 7.56 & 11.6 $\pm$ 11.5 & 0.40 & 19.06 & 11.3 $\pm$ 11.5 & 0.40 & 3.52 \\
& \gclr 035 power drill & \gclr 0.6 $\pm$ 1.3 & \gclr 0.87 & \gclr 10.04 & \gclr 4.0 $\pm$ 6.8 & \gclr 0.70 & \gclr 15.57 & \gclr 11.0 $\pm$ 12.8 & \gclr 0.43 & \gclr 1.63 \\ \midrule
& Average & 9.3 $\pm$ 7.4 & 0.50 & 7.02 & 9.6 $\pm$ 9.8 & 0.49 & 13.69 & 8.6 $\pm$ 10.9 & 0.56 & 2.02 \\ \midrule
& & & & & & & & & & \\

\parbox[t]{2mm}{\multirow{3}{*}{\rotatebox[origin=c]{90}{\scriptsize{Test set 6}}}} 
& \gclr 005 tomato soup can & \gclr 11.4 $\pm$ 10.5 & \gclr 0.32 & \gclr 4.17 & \gclr 11.5 $\pm$ 9.8 & \gclr 0.32 & \gclr 10.21 & \gclr 9.1 $\pm$ 12.0 & \gclr 0.60 & \gclr 2.10 \\
& 037 scissors & 9.9 $\pm$ 9.9 & 0.39 & 2.55 & 11.3 $\pm$ 9.2 & 0.31 & 4.11 & 18.3 $\pm$ 10.4 & 0.19 & 0.71 \\
& \gclr 040 large marker & \gclr 19.6 $\pm$ 5.3 & \gclr 0.03 & \gclr 1.38 & \gclr 20.0 $\pm$ 6.4 & \gclr 0.07 & \gclr 2.95 & \gclr 18.2 $\pm$ 10.2 & \gclr 0.19 & \gclr 0.51 \\  \midrule
& Average & 13.6 $\pm$ 8.6 & 0.25 & 2.70 & 14.3 $\pm$ 8.4 & 0.23 & 5.76 & 15.2 $\pm$ 10.9 & 0.33 & 1.11 \\ \bottomrule

\end{tabular}}
\end{center}
\caption{\textbf{Generalization to Unseen Objects}. We evaluate generalization to unseen objects and compare our model with the baselines. We create six different test sets of three objects, which we leave out during training. We report the detailed results per test set in this table. }
\label{tab:app_generalization}
\end{table*}

\begin{table}[t]
\centering
\resizebox{0.8\columnwidth}{!}{
\begin{tabular}{cl}
\textbf{Notation} & \textbf{Meaning} \vspace{0.1cm}\\
\hline\\
$\statevec$ & state \\
$\actions$ & action \\
$\gpolicyvec$ & grasping policy \\
$\mpolicyvec$ & motion synthesis policy \\
$\grasplabel$ & static grasp label \\
$\posvec$ & 3D joint position \\
$\mathbf{q}$& joint angles \\
$\posesix$& 6D pose \\
& \\
$\posesix_h$ & 6D global hand pose \\
$\dot{\posesix}_h$ & 6D global hand velocities \\
$\posesix_o$ & 6D object pose \\
$\dot{\posesix}_o$ & 6D object velocities \\
$\posesix_g$ & 6D goal object pose \\
$\posevec$ & hand joint angles\\
$\dot{\posevecgen}_h$ & hand joint angular velocities\\
$\overline{\posesix}_h$ & 6D global hand pose in grasp label \\
$\overline{\posesix}_o$ & 6D global object pose in grasp label \\
$\overline{\mathbf{q}}_h$ & 3D hand pose in grasp label\\
$\overline{\mathbf{g}}_c$ & target contacts\\

$\overline{\posvec}$ & 3D target joint position \\
$\forcevec$ & contact forces \\
& \\
$\torques$ & joint torques \\
$k_p$ & PD-controller parameter \\
$k_d$ & PD-controller parameter \\
$\posevecgen_{\text{ref}}$ & reference joint angles \\
$\posevecgen_b$ & bias joint angle term \\
& \\
$\phi(\cdot)$ & feature extractor \\
$\widetilde{\cdot}$ & transformation to wrist reference frame \\
$\widetilde{\posesix}_o$ & 6D object pose in wrist reference frame \\
$\dot{\widetilde{\posesix}}_o$ & 6D object velocities in wrist reference frame \\
$\dot{\widetilde{\posesix}}_h$ & 6D global hand velocities in wrist reference frame \\
$\widetilde{\posvec}_z$ & vertical distance to surface where object rests \\

\end{tabular}
}
\captionof{table}{Glossary (part 1) for the notation used in this paper.}
\label{tab:glossary_p1}
\end{table}

\begin{table}[t]
\centering
\resizebox{1.0\columnwidth}{!}{
\begin{tabular}{cl}
\textbf{Notation} & \textbf{Meaning} \vspace{0.1cm}\\
\hline\\

& \\
$\goals$ & goals \\
$\widetilde{\goalvec}_x$ & 3D distance between current and target joint positions \\
$\widetilde{\goalvec}_q$ & angular distance between current and target joint/wrist rotations \\
$\goalvec_c$ & contact vector \\
$\goalvec_{o,x}$ & 3D distance between current and target object position \\
$\goalvec_{o,q}$ & angular distance between current and target object rotation \\
& \\
$r$ & total reward for grasping \\
$r_x$ & position reward \\
$w_{x}$ & position reward weight \\
$r_q$ & pose reward \\
$w_{q}$ & pose reward weight \\
$r_c$ & contact reward \\
$w_{c}$ & contact reward weight \\
$\lambda$ & contact reward coefficient \\
$m_\text{o}$ & object's mass \\
$r_\text{reg}$ & regularizing reward term \\
$w_{\text{reg},h}$ & regularizing reward term hand weight \\
$w_{\text{reg},o}$ & regularizing reward term object weight \\
& \\
$r_m$ & total reward motion synthesis\\
$r_{m,x}$& position reward motion synthesis\\
$r_{m,q}$& pose reward motion synthesis\\
$\alpha_{x}$ & position reward weight motion synthesis\\
$\alpha_{q}$ & pose reward weight motion synthesis\\
& \\
$\psi(\cdot)$ & feature extractor motion synthesis \\
$\widehat{\mathbf{T}}_h$ & estimated 6D target hand pose \\
$\mathbf{T}_\text{pd}$ & 6D pose input to the PD-controller for motion synthesis \\
& \\
$\overline{\mathbf{v}}_h$ & hand mesh \\
$\overline{\vertexvec}_{o}$ & object mesh\\
& \\

\end{tabular}
}
\captionof{table}{Glossary (part 2) for the notation used in this paper.}
\label{tab:glossary_p2}
\end{table}



\end{document}